\documentclass[10pt,twocolumn,letterpaper]{article}

\usepackage{times}
\usepackage{epsfig}
\usepackage{graphicx}
\usepackage{amsmath}
\usepackage{amssymb}
\usepackage{kotex}
\usepackage{booktabs}
\usepackage{array}
\usepackage{multirow}
\usepackage{paralist}
\usepackage{caption}
\usepackage{diagbox}
\usepackage[dvipsnames]{xcolor}

\newcolumntype{k}{>{\centering\arraybackslash}p{2.7em}}

\newcommand\pp{\raisebox{0.5ex}{\scriptsize{++} }}

\newcommand\cn[1]{{\color{ForestGreen}#1}}
\newcommand\st[1]{{\color{Orange}#1}}

\usepackage{authblk}

\renewcommand*{\Affilfont}{\normalsize}
\makeatletter
\renewcommand\AB@affilsepx{\quad \protect\Affilfont}
\makeatother

\usepackage{cvpr}              

\definecolor{cvprblue}{rgb}{0.21,0.49,0.74}
\usepackage[pagebackref,breaklinks,colorlinks,citecolor=cvprblue]{hyperref}

\def\paperID{7068} 
\def\confName{CVPR}
\def\confYear{2024}

\title{MoST: Motion Style Transformer between Diverse Action Contents}

\author[1,2,3]{Boeun Kim}
\author[1]{Jungho Kim}
\author[3]{Hyung Jin Chang}
\author[2]{Jin Young Choi}
\affil[1]{Korea Electronics Technology Institute}
\affil[2]{Seoul National University}
\affil[3]{University of Birmingham}

\begin{document}
\maketitle
\begin{abstract}
While existing motion style transfer methods are effective between two motions with identical content, their performance significantly diminishes when transferring style between motions with different contents. This challenge lies in the lack of clear separation between content and style of a motion. To tackle this challenge, we propose a novel motion style transformer that effectively disentangles style from content and generates a plausible motion with transferred style from a source motion. Our distinctive approach to achieving the goal of disentanglement is twofold: (1) a new architecture for motion style transformer with `part-attentive style modulator across body parts' and `Siamese encoders that encode style and content features separately'; (2) style disentanglement loss. Our method outperforms existing methods and demonstrates exceptionally high quality, particularly in motion pairs with different contents, without the need for heuristic post-processing.
Codes are available at \href{https://github.com/Boeun-Kim/MoST}{\textit{\color{magenta}https://github.com/Boeun-Kim/MoST}}.
\end{abstract}
\section{Introduction}

\begin{figure}[t]
\centering
    \includegraphics[width=\linewidth]{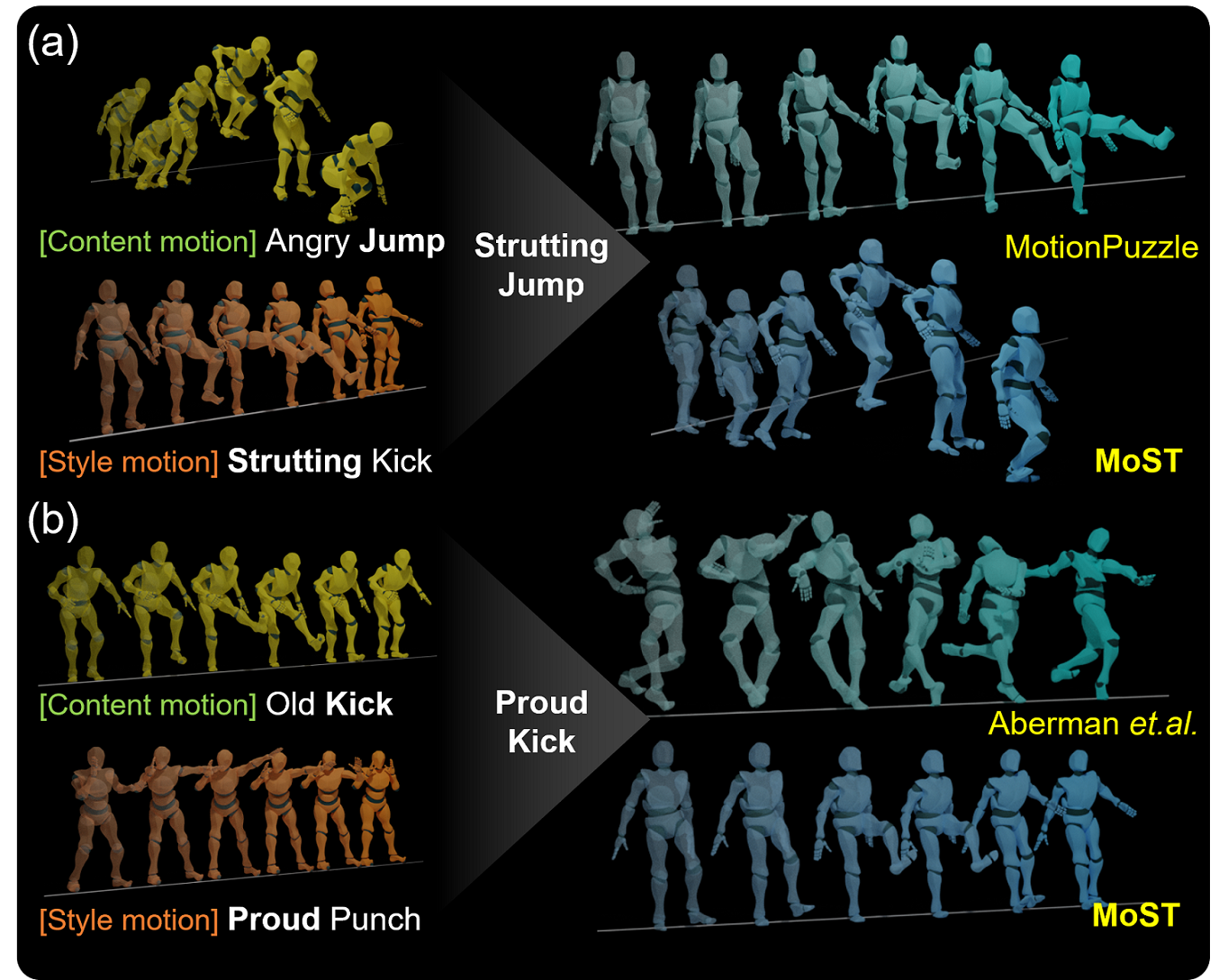}
    \caption{Frequent failure cases in existing methods: (a) A result of MotionPuzzle~\cite{jang2022motion} replicating \textit{style motion}. (b) A result of Aberman \etal~\cite{aberman2020unpaired} showing complete failure with twisted motion. The character for the visualization is sourced from Mixamo~\cite{mixamo}.
    }
    \label{fig:limitations}
\vspace{-0.3cm}
\end{figure}

Motion style transfer has received considerable attention for its potential to generate character animations reflecting human personalized characteristics in applications such as games, metaverse, and robotics.
Our goal is to transfer stylistic characteristics from a source motion sequence (\textit{style motion}) to a target motion sequence (\textit{content motion}) without manually providing style labels.
Performing this style transfer between motion sequences is a considerably more challenging task~\cite{aberman2020unpaired,park2021diverse,wen2021autoregressive, jang2022motion} compared to generating motion when both content and style labels are manually provided~\cite{tao2022style, chang2022unifying}.

The most significant challenge is transfer failure, where the generated motion loses \cn{content} of the \textit{content motion} or does not reflect \st{style} of the \textit{style motion}.
Transfer failure often occurs when the \cn{contents} of the \textit{style motion} and \textit{content motion} differ from each other, \textit{e.g.} kick and punch.
Certainly, its complexity arises from the significant variation in style expression between different parts of the body for distinct \cn{contents}. 
Despite the practical importance of transferring \st{styles} between different \cn{content}, existing approaches have not primarily addressed these cases, nor have they been comprehensively evaluated, either qualitatively or quantitatively.
Fig.~\ref{fig:limitations} illustrates representative failure cases in existing methods.
They fail to clearly disentangle \st{style} from source \cn{content} and appropriately transfer it to the target \cn{content}.
Fig.~\ref{fig:limitations} (a) illustrates the result of MotionPuzzle~\cite{jang2022motion}, where the generated motion, intended to be a jump, instead displays a kick. This method tends to replicate the motion pattern of the \textit{style motion} rather than effectively transferring disentangled \st{style}.
In Fig.~\ref{fig:limitations} (b), Aberman \etal~\cite{aberman2020unpaired} utterly fails to generate a plausible motion.
Aberman \etal recognized this issue, acknowledging that their method is less effective in motions other than walking and running.

The additional notable limitation found in recent studies~\cite{aberman2020unpaired,park2021diverse,wen2021autoregressive, jang2022motion} is the requirement for heavy post-processing.
These methods generate only the body pose of each frame and cannot generate the global translation. 
Instead, they copy the global translation from the \textit{content motion} in a post-processing step.
However, this copying process degrades the quality of the generated motion, as highlighted by Park \etal~\cite{park2021diverse} and Jang \etal~\cite{jang2022motion} in their papers.
Park \etal point out that inconsistencies may be caused between the generated pose and the copied global translation.
Furthermore, methods of \cite{aberman2020unpaired, park2021diverse, jang2022motion} heuristically grab the foot contact timings in the \textit{content motion} and forcibly fix the generated motion to make the feet touch the ground at those specific timings.
Without fixing, feet often penetrate the floor or float in the air in their generated motions.

To address the primary concern of transfer failure, we design a new framework called MoST and new loss functions for effectively transferring \st{style} between different \cn{contents}.
As \st{style} manifests in different patterns across various body parts based on \cn{content}, MoST handles body part-specific style and content features.
MoST comprises transformer-based Siamese motion encoders, a part-attentive style modulator (PSM), and a motion generator.
In contrast to the previous methods~\cite{aberman2020unpaired, park2021diverse, jang2022motion} that build two separate networks for encoding content and style features, we introduce Siamese motion encoders capable of simultaneously extracting both features from an input motion.
Subsequently, PSM modulates the raw style feature extracted from the \textit{style motion} to align with the \cn{content} of the \textit{content motion} before being inserted into the generator. 
This modulation enables effective disentanglement of \st{style} from the \textit{style motion} and its expression in the \textit{content motion}.
Furthermore, we propose new loss functions.
The style disentanglemfent loss aims to distinctly separate \st{style} and \cn{content}. This separation enhances the model's robustness in transferring \st{style}, irrespective of the \cn{content} of the \textit{style motion}.

Our resulting motions are of high quality, eliminating the need for any heuristic post-processing.
MoST achieves this by generating global translational motion along with pose sequences.
In the proposed framework, the global translational feature and body part features exchange their information via attention, ensuring consistency between the generated poses and global translation.
Moreover, to mitigate issues such as jittering and foot-skating, we introduce a physics-based loss. 
We evaluated the proposed model on two representative motion capture datasets~\cite{xia2015realtime,aberman2020unpaired}, verifying its performance both qualitatively and quantitatively.
The main contributions of this study are summarized as follows:
\begin{compactitem}
    \item 
    We design MoST, incorporating Siamese encoders and PSM, to effectively disentangle \st{style} from the source motion and align it with the target \cn{content}.
    \item We propose novel loss functions to improve the model's ability to disentangle \st{style} from \cn{content} within the motion and generate plausible motion.
    \item 
    MoST substantially outperforms existing methods, particularly when the two inputs have different contents.
    Furthermore, our method achieves high fidelity in its output without the need for any post-processing.
\end{compactitem}

\begin{figure*}[t]
\centering
    \includegraphics[width=\linewidth]{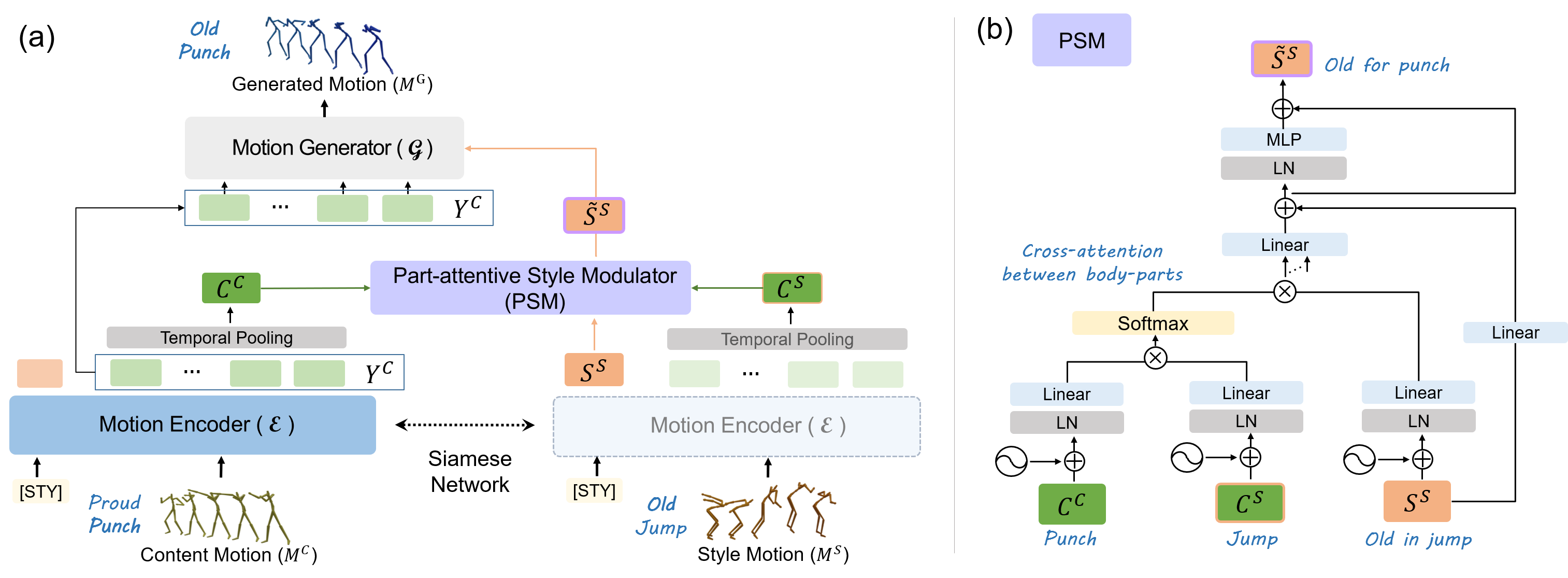}
    \caption{(a) Overall framework of MoST comprising Siamese motion encoders $\mathcal{E}$, motion generator $\mathcal{G}$, and part-attentive style modulator (PSM). PSM modulates style feature $S^S$ under the condition of both contents of content motion and style motion, \textit{i.e.}, $C^C$ and $C^S$. $\mathcal{G}$ generates final output motion with content dynamics feature $Y^C$ and the modulated style feature $\tilde{S}^S$.
    (b) Detailed operations in PSM
    }
    \label{fig:framework}
\end{figure*}

\section{Related Works}
\subsection{Motion and Image Style Transfer}
Style transfer methods usually have been developed in the image domain first~\cite{gatys2016image,johnson2016perceptual,huang2017arbitrary, kingma2018glow,choi2020stargan} and applied to the motion domain~\cite{holden2016deep,holden2017phase,du2019stylistic,aberman2020unpaired,wen2021autoregressive,park2021diverse}.
In image style transfer, Gatys \etal~\cite{gatys2016image} first obtained style and content features from a neural network rather than hand-crafted features. The style transfer model is optimized to yield features similar to both style and content features extracted from the given input images.
Johnson \etal~\cite{johnson2016perceptual} proposed a real-time optimization framework by using the perceptual loss.
The work of Holden \etal~\cite{holden2016deep} was the first in motion style transfer, which adopts the optimization framework of Gatys \etal~\cite{gatys2016image}.
Holden \etal~\cite{holden2017phase} and Du \etal~\cite{du2019stylistic} improved run-time and memory efficiency by replacing a training procedure similar to the approach of Johnson \etal~\cite{johnson2016perceptual}.

Recently, Huang \etal~\cite{huang2017arbitrary} introduced AdaIN, which simply replaces style statistics in certain layers during training. 
Inspired by the work of Huang \etal~\cite{huang2017arbitrary}, Aberman \etal~\cite{aberman2020unpaired} proposed temporally invariant AdaIN which is applied in the temporal convolution layers.
Kingma \etal~\cite{kingma2016improved,kingma2018glow} first introduced a generative flow~\cite{dinh2014nice} into an image generation framework.
Few years later, Wen \etal~\cite{wen2021autoregressive} adopted the generative flow model into motion style transfer, generating plausible motions with less jittering.
However, the method is limited in capturing the global style feature owing to the extraction of the style feature from the small window.
Furthermore, the model has large parameters and is extremely slower than other methods~\cite{aberman2020unpaired,park2021diverse}.
Choi \etal~\cite{choi2020stargan} proposed an image-to-image translation method aimed at multiple domains. 
Park \etal~\cite{park2021diverse} adopted this approach regarding the styles as domains.
The style encoder successfully separates distinctive style features in the latent space due to the head layers, which are constructed separately for each style.
However, a style label should be injected into the network along with the style motion sequence in this method.
Furthermore, pose twisting and jittering are frequently shown in the generated motion.
The recent methods of motion style transfer~\cite{aberman2020unpaired,wen2021autoregressive,park2021diverse} have common shortcomings.
The methods encounter difficulties in transferring styles between motions with different contents, and they require hard post-processing, such as fixing foot contact or copying global translation.

Most recently, StyleFormer~\cite{wu2021styleformer} and StyTr$^2$~\cite{deng2022stytr2} have successfully applied the transformer architecture to image style transfer, generating stylized images through cross-attention between content and style features.
Taking inspiration from these works, we introduce the transformer architecture to the motion encoder and generator. 
Motion requires more precise modeling than images because even small temporal inconsistencies or incorrect joint movements are noticeable. Therefore, eliminating the original style and injecting a new style into the motion demands careful consideration. Our framework aims to disentangle content and style clearly and modulate style to be expressed effectively in the target motion, resulting in a plausible output motion.

\subsection{Recent Studies on Motion Modeling}
There are some methods that synthesize motion using given labels of style and content~\cite{tao2022style, chang2022unifying}.
Chang \etal introduced a diffusion-based model, and Tao \etal proposed an online synthesis method.
Tang \etal introduced a method to stylize transitional motion between two frames~\cite{tang2023rsmt}.
Our goal is more complex compared to these methods, as we aim to transfer style between two input motions of diverse action types rather than using labels directly.
Recently, there have been many studies that generate motion from a textual description~\cite{petrovich2022temos,tevet2022motionclip,chen2023executing,dabral2023mofusion}.
MotionCLIP~\cite{tevet2022motionclip} generates motions from arbitrary texts owing to the alignment of a human motion manifold and contrastive language-image pre-training
(CLIP)~\cite{radford2021learning} space.
MLD~\cite{chen2023executing} and MoFusion~\cite{dabral2023mofusion} generate realistic motions from long-sequence descriptions utilizing conditional diffusion models.

Similar to our task, MotionPuzzle~\cite{jang2022motion} presents an innovative framework that transfers manually specified motion into a desired body part, allowing the combination of multiple target motions.
However, the method has limitations in disentangling the style and content characteristics, resulting in the duplication of the \textit{style motion} instead of transferring only the disentangled style.
Nevertheless, MotionPuzzle provided essential inspiration for our work handling motion features by body parts for precise motion modeling.

\section{Method}
\subsection{Overall Framework}
The proposed framework aims to transfer the \st{style} from a given \textit{style motion} $M^S$ to a given \textit{content motion} $M^C$ to generated $M^G=\mathtt{MoST}(M^C, M^S)$.
As depicted in Fig.~\ref{fig:framework} (a), our framework comprises Siamese motion encoders $\mathcal{E}$, motion generator $\mathcal{G}$, and a part-attentive style modulator (PSM).
$\mathcal{E}$ is designed to encode both a content feature and a style feature simultaneously from a single motion. 
We utilize $\mathcal{E}$ to encode both $M^C$ and $M^S$ as
$\mathcal{E}(M^C) = \{S^C, Y^C\}$, $\mathcal{E}(M^S) = \{S^S, Y^S\}$, where
$Y^{(\cdot)}$ denotes frame-level content dynamics features and $S^{(\cdot)}$ denotes a body part-specific style feature.
The content feature of $M^S$, which is unobtainable by the style encoders in previous methods, is utilized as additional information for subsequent steps.
Subsequently, $S^S$ is modulated through the PSM as $\mathtt{PSM}(S^S | C^S, C^C) = \tilde{S}^S$, where $C^C$ and $C^S$ are obtained by temporally pooling $Y^C$ and $Y^S$.
Finally, $\mathcal{G}$ produces the generated motion $M^G$ from $Y^C$ based on the modulated style feature $\tilde{S}^S$ as $\mathcal{G}(Y^C, \tilde{S}^S)= M^G$.

\subsection{Motion Representation}
\label{sec:motion_representation}
The motion sequence is defined by $M = \{m_t^j,m_t^{root},$ $ v_t| j=1,\cdots,J;~ t=1,\cdots,T\}$, where $J$ is the number of body joints and $T$ is the maximum length of the model input.
$\{m_t^j\in \mathbb{R}^{7}|j=1,\cdots,J\}$ denotes local motion, where $m_t^j$ represents the $j$-th joint vector in the $t$-th frame.
$m_t^{root}\in \mathbb{R}^{7}$ and $v_t\in \mathbb{R}^{4}$ denote global translation and global velocity, respectively. Note that $M^C$, $M^S$, and $M^G$ have the same form as $M$.
$m_t^j$, $m_t^{root}$, and $v_t$ contain both positional and angular representations.
More details are available in the supplementary material.

We generate motion embeddings that are the input tensors of $\mathcal{E}$.
Inspired by the method of~\cite{jang2022motion}, we grouped the joints into $P$ body parts, accounting for the body structure.
Moreover, to consider global translational motion, which existing methods overlook, we acquire a separate embedding of global translation in addition to body part embeddings.
$i$-th body part embedding at $t$-th frame, is written as
\begin{align}
    \bar{p}_{(t,i)} =\mathtt{FC}(p_{(t,i)})\in \mathbb {R}^{d}, ~
    p_{(t,i)} = ||_{j}^{J^i}m_t^{j} \in \mathbb{R}^{7N_{J^i}},
\end{align}
where $p_{(t,i)}$ is a concatenated vector of local joint vectors in the $i$-th body part.
$J^i$ is the set of joints within the $i$-th part, and $N_{J^i}$ denotes the number of joints in $J^i$. 
$d$ denotes the feature dimension of each body part embedding.
$\mathtt{FC}(\cdot)$ and $||$ indicate the fully-connected layer and the concatenation operation, respectively.
The global translation embedding $\bar{g}_t$ is obtained as
\begin{align}
\begin{split}
    \bar{g}_t &= [\bar{m}_t^{root} ; \bar{v}_t] \in \mathbb {R}^{d}, \\
    \bar{m}_t^{root} = \mathtt{FC}(m_t^{root})&\in \mathbb {R}^{d/2}, \quad
    \bar{v}_t =\mathtt{FC}(v_t )\in \mathbb {R}^{d/2}.
\end{split}
\end{align}
Subsequently, the final embedding tensor $X$ is expressed as
\begin{align}
    X &= [X_1, \cdots, X_t, \cdots, X_T]^\top \in \mathbb{R}^{T \times (P+1) \times d}, \\
    X_t &= [\bar{p}_{(t,1)}, \cdots, \bar{p}_{(t,i)}, \cdots, \bar{p}_{(t,P)}, \bar{g}_t]^\top \in \mathbb{R}^{(P+1) \times d}. \notag
\end{align}

\subsection{Siamese Motion Encoders}
The frameworks of existing methods~\cite{aberman2020unpaired,park2021diverse,jang2022motion} consist of separate style encoders and content encoders. 
However, we introduce a Siamese architecture for both encoders to eliminate redundancy. Furthermore, we design the encoder to extract both style and content features from each motion, which are utilized in the next step.
In contrast to the style encoders of Park \etal~\cite{park2021diverse} and Wen \etal~\cite{wen2021autoregressive}, which can only extract style from a short-range time window, our model extracts a global style feature across the entire motion sequence.

Motion encoder, $\mathcal{E}(M^{(\cdot)}) = \{S^{(\cdot)}, Y^{(\cdot)}\}$, comprises $N$ stacked transformer blocks, and each block contains body part attention and temporal attention modules.
We adopt the transformer architecture from the method of~\cite{kim2022global}, but we employ part attention instead of joint attention.
Additionally, we introduce a style token to aggregate the style features across an entire motion sequence.
Let us denote $Z^{n-1}$ as the input motion tensor of the $n$-th block.
The input tensor of the first block is represented as 
\begin{align}
    Z^0 &= [X_{sty}, X_1, \cdots, X_T]^\top \in \mathbb{R}^{(T+1) \times (P+1) \times d},
\end{align}
where $X_{sty}\in \mathbb{R}^{(P+1) \times d}$ is a trainable style token.
A positional embedding tensor is defined as $E = [E_1, \cdots, E_t, \cdots, E_{(T+1)}]^\top \in \mathbb{R}^{(T+1) \times (P+1) \times d}$~\cite{vaswani2017attention,kim2022global}.
In each block, the part attention module learns dependencies between body parts at the $t$-th frame as
\begin{align}
    \label{eq:sp-att}
    \bar{Z}_t^{n-1} = \mathtt{MHA}(\mathtt{LN}(Z_t^{n-1}+E_t)) + (Z_t^{n-1}+E_t),
\end{align}
where $Z_t^{n-1}\in \mathbb{R}^{(P+1) \times d}$ is a sliced vector of $Z^{n-1}$.
$\mathtt{LN}(\cdot)$ denotes the layer normalization operator and
$\mathtt{MHA}(\cdot)$ denotes the multi-head attention operator.
Subsequently, the temporal attention module learns the dependencies between frames as
\begin{align}
    \label{eq:temp-att}
    Z^{n} = \mathtt{MHA}(\mathtt{LN}(\bar{Z}^{n-1}+E)) + (\bar{Z}^{n-1}+E),
\end{align}
where $\bar{Z}^{n-1} = [\bar{z}_{sty}^{n-1}, \cdots, \bar{z}_T^{n-1}]^\top \in \mathbb{R}^{(T+1) \times (P+1)d}$ and $\bar{z}_t^{n-1} \in \mathbb{R}^{(P+1)d}$ is a vectorized form of $\bar{Z}_t^{n-1}$.
We extract the style feature matrix from the output of the $(N-1)$-th block as
\begin{align}
    S^{(\cdot)} = Z_{sty}^{N-1}  \in \mathbb{R}^{(P+1) \times d}.
\end{align}
To obtain a content feature, $\hat{Z}^{N-1} = [Z_1^{N-1}, \cdots, Z_T^{N-1}]^\top$ go through the last transformer block.
In the last block, instance normalization (IN) is applied to remove style characteristics from the motion features~\cite{huang2017arbitrary, aberman2020unpaired,park2021diverse} as
\begin{align}
    \check{Z}^{N-1}_{t,i} = \frac{\hat{Z}^{N-1}_{t,i} -\mu(\hat{Z}^{N-1})}{\sigma(\hat{Z}^{N-1})},
\end{align}
where $\mu(\cdot)$ and $\sigma(\cdot)$ are operations that yield $d$-dimensional channel-wise mean and variance vectors from a tensor.
$\check{Z}^{N-1}$ passes through body part attention and temporal attention same as $Z^{n-1}$ in Eqs.(\ref{eq:sp-att}) and (\ref{eq:temp-att}).
Subsequently, the frame-level content dynamics feature is encoded as
\begin{align}
    Y^{(\cdot)} &= [\check{Z}_1^N, \cdots, \check{Z}_T^N]^\top  \in \mathbb{R}^{T \times (P+1) \times d}.
\end{align}

\subsection{Part-Attentive Style Modulator}
Part-attentive style modulator (PSM) modulates the style feature $S^S$, which originates from $M^S$, to be more effectively expressed in $M^C$.
The detailed operation is illustrated in Fig.~\ref{fig:framework} (b). 
Initially, $S^S$ is expressed within the context of $C^S$, and our objective is to derive $\tilde{S}^S$, suitable for integration into $C^C$.
We initiate the process by computing cross-attention between two content features, $C^S$ and $C^C$, where $C^{(\cdot)} \in \mathbb{R}^{(P+1) \times d}$ is obtained by temporally pooling $Y^{(\cdot)}$.
This cross-attention identifies how \st{style} should be transmitted from a specific body part of $C^S$ to a corresponding body part of $C^C$, \textit{e.g.} the 
legs in jump to the arm in punch.
Subsequently, $\tilde{S}^S$ is generated by linear combinations of part-specific style features of $S^S \in \mathbb{R}^{(P+1) \times d}$ according to the cross-attention.
Consequently, PSM prevents the transmission of motion from an undesired body part of $M^S$, \ie it only transfers \st{style} well-disentangled from \cn{content} in $M^S$. As a result, the generated motion preserves the distinct \cn{content} of $M^C$ instead of confused combinations of motions from $M^S$ and $M^C$.
The cross-attention mechanism is applied as
\begin{align}
    &\mathtt{cross\text{-}MHA}(\bar{C}^C, \bar{C}^S, \bar{S}^S) = (||_{i=1}^h H_i)W_H, \\ 
    H_i =& \mathtt{Attn}(\mathtt{LN}(\bar{C}^C) W_i^Q, \; \mathtt{LN}(\bar{C}^S) W_i^K, \; \mathtt{LN}(\bar{S}^S) W_i^V), \notag \\ 
    \bar{C}^C &= {C}^C + E_p, \; \bar{C}^S = {C}^S + E_p, \; \bar{S}^S = {S}^S + E_p,\notag
\end{align}
where $\mathtt{Attn}(Q,K,V)$ indicates $\mathtt{softmax}(QK^\top/\sqrt{d})V$. $W^Q, W^K, W^V \in \mathbb{R}^{d \times d'}$ and $W_H \in \mathbb{R}^{d'h \times d}$ are weight matrices for projection. $d'$ and $h$ denote the projection dimension and the number of heads, respectively.
As the index indicating the body part can serve as crucial information, we incorporate positional embedding $E_p \in \mathbb{R}^{(P+1) \times d}$ for each input.
Finally, $\bar{S}^S$ is calculated as
\begin{align}
    \check{S}^S &= \mathtt{FC}(S^S) + \mathtt{cross\text{-}MHA}(\bar{C}^C, \bar{C}^S, \bar{S}^S),\\
    \tilde{S}^S &= \check{S}^S + \mathtt{MLP}(\mathtt{LN}(\check{S}^S)),
\end{align}
where $\mathtt{MLP}(\cdot)$ denotes multi-layer perceptron. 

\subsection{Motion Generator}
The motion generator, denoted as $\mathcal{G}(Y^C, \tilde{S}^S) = M^G$, is composed of $N$ stacked transformer blocks, similar to $\mathcal{E}$.
$Y^C$ is introduced as an input tensor for the first transformer block, denoted as $U^0 = Y^C$, where $U^{n-1}$ represents the input tensor of the $n$-th block.
Since $Y^C$ encapsulates extensive information regarding the frame-level dynamics of each body part, \cn{content} can be reliably conveyed to $M^G$.
To incorporate $\tilde{S}^S$, we employ AdaIN~\cite{huang2017arbitrary}, a method proven to be effective in previous works~\cite{aberman2020unpaired,park2021diverse}.
In each transformer block, AdaIN is applied as
\begin{align}
\begin{split}
    \hat{U}^{n-1}_{t,i} &= \gamma \left(\frac{U^{n-1}_{t,i} -\mu(U^{n-1})}{\sigma(U^{n-1})}\right) + \beta, \\
    \gamma = \mathtt{FC}&(\tilde{S}^S) \in \mathbb{R}^d, \quad
    \beta = \mathtt{FC}(\tilde{S}^S)\in \mathbb{R}^d,
\end{split}
\end{align}
where $\beta$ and $\gamma$ are parameters modifying the channel-wise mean and variance, respectively.
$\hat{U}^{n-1}$ passes through part attention and temporal attention consistent with $Z^{n-1}$ in Eqs.(\ref{eq:sp-att}) and (\ref{eq:temp-att}). 
The local motion, global translation, and global velocity information of the generated motion are reconstructed by the output feature tensor of the final transformer block, $\hat{U}^N$, as
\begin{align}
    \{\hat{m}_t^j \in \mathbb{R}^{7}|j \in {J^i}\} &= \mathtt{FC}(\hat{U}^N_{t,i}), \;\; (i \leq P), \label{gen_motions}\\
    \hat{m}_{t}^{root} &= \mathtt{FC}(\hat{U}^N_{t,(P+1)}) \in \mathbb{R}^{7}, \\
    \hat{v}_{t} &= \mathtt{FC}(\hat{U}^N_{t,(P+1)}) \in \mathbb{R}^{4}.
\end{align}
Finally, the generated motion becomes \\
$$M_G = \{\hat{m}_t^j,\hat{m}_t^{root}, \hat{v}_t| j=1,\cdots,J;~t=1,\cdots,T\}.$$

\subsection{Loss}
We introduce the \textit{style disentanglement loss}, aiming to effectively separate \st{style} from \cn{content}.
It increases the robustness of our model in generating a well-stylized motion regardless of \cn{content} of $M^S$.
It is written as
\begin{align}
\label{eq:d}
    L_{D} = \mathbb{E}_{M^C, M^S_{a}, M^S_{b} \sim \mathbb{M}} \; &||\; \mathtt{MoST}(M^C, M^S_{a})\\ \notag
    &-\mathtt{MoST}(M^C, M^S_{b}) ||_2,
\end{align}
where $\mathbb{M}$ denotes the training set. We denote $||\{\cdot\}||_2$ as a sum of $l_2$-norms of all vectors in set $\{\cdot\}$ for simplicity.
$L_{D}$ minimizes the discrepancy between the generated motions stylized by $M^S_{a}$ and $M^S_{b}$, where $M^S_{a}$ and $M^S_{b}$ denote two \textit{style motions} that have identical style labels but different content labels.
$L_{D}$ induces the model to clearly remove \cn{content} from $M^S$, independent of the specific \cn{content} present in $M^S$, thereby avoiding the blending of \cn{content} into $M^G$.

\newcolumntype{?}{!{\vrule width 2pt}}
\begin{figure*}[t]
\begin{minipage}{0.36\linewidth}
    \centering
    \includegraphics[width=\linewidth]{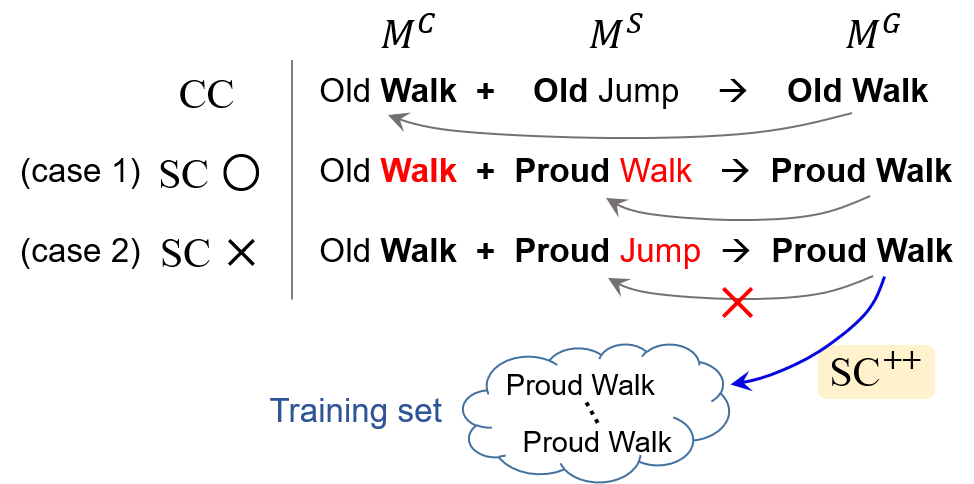}
    \caption{Description of evaluation metrics, using easy-to-recognize label notations. Note that our model uses only motion data}
    \label{fig:metric}
\end{minipage}
\hspace{0.2cm}
\begin{minipage}{0.62\linewidth}
    \captionof{table}{Motion style transfer results evaluated in content consistency (CC) and style consistency\pp (SC\pp) on Xia dataset~\cite{xia2015realtime}. The performances are reported in both cases: when $M^C$ and $M^S$ have identical content labels and different content labels}
    \label{table:sota}
    \centering
    \resizebox{\linewidth}{!}{%
        \begin{tabular}{l|c||c|c||c||c|c}
        \toprule
        \multirow{2}{*}{Methods}& \multicolumn{3}{c||}{CC  $\downarrow$} & \multicolumn{3}{c}{SC\pp $\downarrow$} \\ 
         \cmidrule(lr){2-4} \cmidrule(lr){5-7}
         & \textbf{average} & same cnt.  & diff. cnt. & \textbf{average} & same cnt. & diff. cnt. \\ \midrule
        MotionPuzzle~\cite{jang2022motion}         & 51.4 & 8.6 & 66.9  & 76.0 & 58.0 & 82.5 \\
        Aberman \etal~\cite{aberman2020unpaired}    & 46.0 & 42.0 & 47.0  & 189.7 & 220.8 & 178.4 \\
        Park \etal~\cite{park2021diverse}           & 38.4 & 22.6 & 44.1  & 65.7 & 59.6 & 67.9 \\
        Wen \etal~\cite{wen2021autoregressive}      & 18.5 & 7.8  & 22.3  & 80.8 & 76.2 & 82.4 \\ \midrule
        MoST                                        & \textbf{8.5} & 8.0 & 8.7  & \textbf{63.0} & 62.2 & 63.2 \\
        \bottomrule
        \end{tabular}%
    }
\end{minipage}
\end{figure*}
Furthermore, we introduce the \textit{physics-based loss} to mitigate pose jittering and improve foot-contact stability. It comprises regularization terms of velocity, acceleration, and foot contact as
\begin{align}
    L_{phy} = \lambda_{vel}R_{vel} +\lambda_{acc}R_{acc}+\lambda_{foot}R_{foot},
\label{eq:phy}
\end{align}
where $\lambda$s denote weights of each regularization term. $R_{vel}$ and $R_{acc}$ are adopted by the method of~\cite{rempe2020contact} and~\cite{jang2022motion}.
Assuming that $\mathtt{MoST}(M^G, M^C)$
should be identical to $M^C$, it should have the same foot contact timing as $M^C$.
Therefore, we introduce foot contact regularization as
\begin{align}
    R_{foot} = \sum_{j}^{J_{feet}}\left(\frac{1}{N_{T_j}}\sum_{t}^{T_{j}}||\dot{\hat{m}}_{t}^j||_2 \right),
\end{align}
where $\dot{\hat{m}}_{t}^j = \hat{m}_{t+1}^j - \hat{m}_{t}^j$. 
$J_{feet}$ includes the toe and heel joints of both feet and $T_{j}$ is a set of frames where the foot joint $j$ contacts the floor in $M^C$. 
$N_{T_j}$ indicates the number of frames in $T_{j}$.
Additionally, commonly used loss functions from existing methods~\cite{aberman2020unpaired, park2021diverse, jang2022motion} are utilized: adversarial ($L_{adv}$), reconstruction ($L_{recon}$), and cycle consistency ($L_{cyc}$) losses.
The final loss is calculated as
\begin{align}
    L =& \lambda_{D} L_{D} + \lambda_{phy} L_{phy} + L_{pre}, \notag\\
    L_{pre} =& \lambda_{adv} L_{adv} + \lambda_{recon} L_{recon} + \lambda_{cyc} L_{cyc}.
    \label{eq:loss}
\end{align}

\section{Experiments}
\label{sec:experiments}
\noindent
\textbf{Implementation Details.}
\\
We evaluate our model using two different motion datasets captured by Xia \etal~\cite{xia2015realtime} (Xia dataset) and Aberman \etal~\cite{aberman2020unpaired} (BFA dataset) following \cite{aberman2020unpaired},~\cite{park2021diverse}, and~\cite{wen2021autoregressive}. 
In both datasets, the skeleton has 31 joints, and we select 21 joints among them for training following\cite{aberman2020unpaired,park2021diverse}.
Xia dataset consists of eight motion styles and six motion contents.
BFA dataset contains 16 styles and has long-length motion sequences not separated by the types of contents. 
The motion sequences are down-sampled into 60fps which is half of the original frame rate. 
The sequence is augmented by applying a random crop during training.
$\mathcal{E}$ and $\mathcal{G}$ comprise three transformer blocks.
We construct a discriminator $\mathcal{D}$ as a single-block transformer for $L_{adv}$.
$T$ is set to 200. We set $\lambda_{adv},\lambda_{D}, \lambda_{vel}, \lambda_{foot}=1$. $\lambda_{recon}, \lambda_{cyc}=3$, and $\lambda_{acc}=0.1$.
Adam optimizer~\cite{kingma2014adam} is used for training, with a learning rate of $e^{-5}$ for $\mathcal{E}$, $\mathcal{G}$, and $e^{-6}$ for $\mathcal{D}$. Our model is trained for 300K iterations with a batch size of 8.
\\
\\
\noindent
\textbf{Evaluation Metrics.}
\\
Motion style transfer methods lack a standardized evaluation metric, hence, we adopt content consistency (CC) and style consistency (SC) proposed by Wen \etal~\cite{wen2021autoregressive}. 
Furthermore, we extend SC to SC\pp to evaluate all the test motion pairs including the cases not covered by SC. 
Fig.~\ref{fig:metric} explains the metrics through straightforward examples.
CC measures how effectively the \cn{content} of $M^C$ is retained in $M^G$.
CC calculates the $l2$-distance between $M^G$ and $M^C$ and averages it across test pairs.
In contrast, SC evaluates how well the \st{style} of $M^S$ is reflected in $M^G$.
SC calculates the similarity between $M^G$ and $M^S$; however, a problem arises when the \cn{content} of $M^S$ differs from that of $M^C$ and $M^G$, as shown in `case 2' of Fig.~\ref{fig:metric}.
We cannot find the reference motion to compare among the input motions.
Therefore, Wen \etal only evaluate cases where the \cn{content} labels of $M^C$ and $M^S$ are identical, \textit{i.e} `case 1'.
To go beyond this limitation and evaluate all motion pairs, in SC\pp, $M^G$ is compared to the pseudo ground truth motions in the training set, which share the identical style and content categories as those of $M^G$.
Formulas are provided in the supplementary material.
A crucial factor to note is that a style transition is deemed successful when both CC and SC\pp are low, as these metrics exhibit a trade-off relationship.
For instance, if a model outputs unmodified $M^C$ as-is to generate $M^G$, SC\pp will yield a high value, although CC may be low.



\newcommand{\onestyle}{\raisebox{-.5\height}{\includegraphics[width=0.30\linewidth]{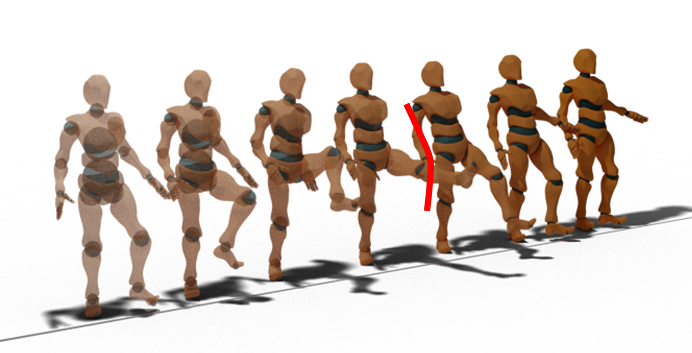}}}
\newcommand{\onecontent}{\raisebox{-.5\height}{\includegraphics[width=0.30\linewidth]{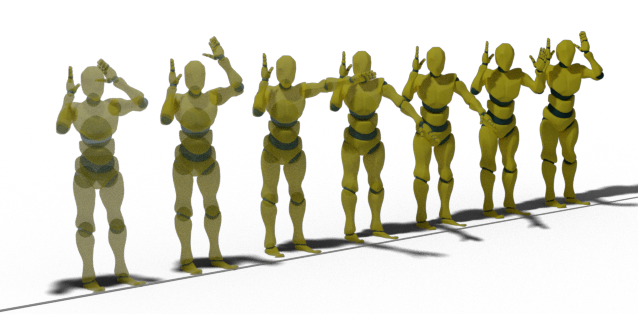}}}
\newcommand{\oneAberman}{\raisebox{-.5\height}{\includegraphics[width=0.30\linewidth]{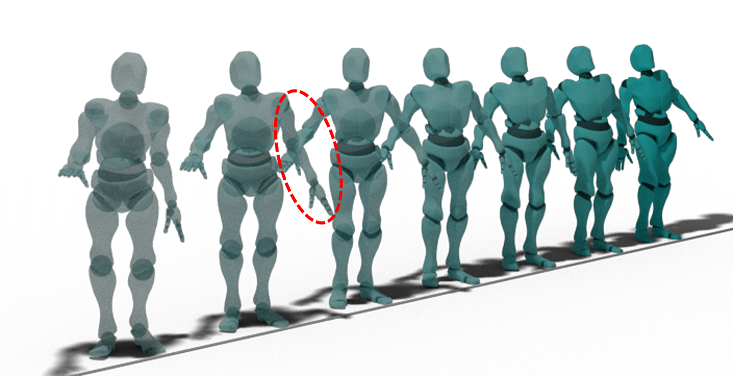}}}
\newcommand{\onePark}{\raisebox{-.5\height}{\includegraphics[width=0.30\linewidth]{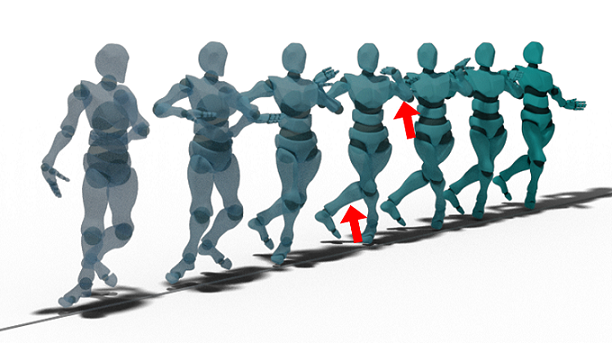}}}
\newcommand{\oneWen}{\raisebox{-.5\height}{\includegraphics[width=0.30\linewidth]{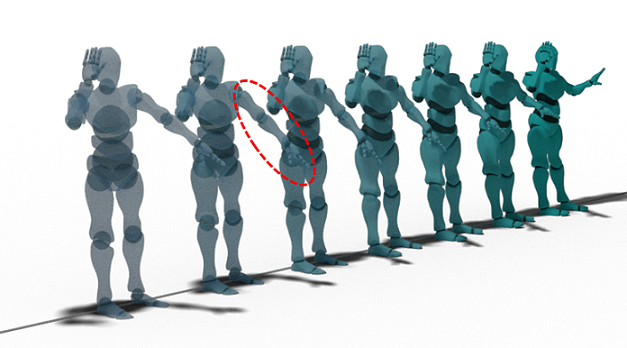}}}
\newcommand{\onePuzzle}{\raisebox{-.5\height}{\includegraphics[width=0.30\linewidth]{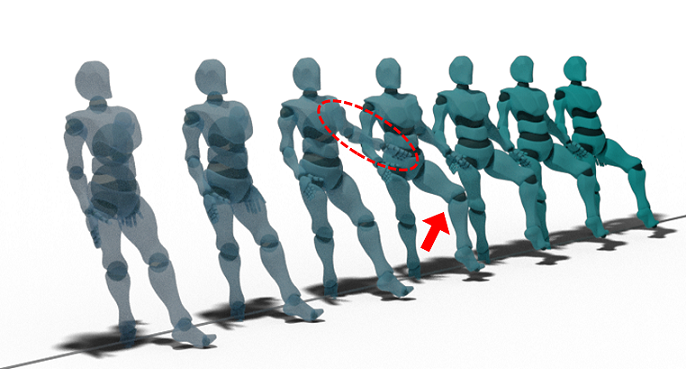}}}
\newcommand{\oneOURS}{\raisebox{-.5\height}{\includegraphics[width=0.30\linewidth]{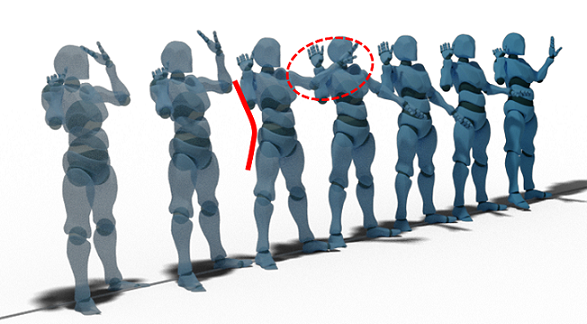}}}

\newcommand{\twostyle}{\raisebox{-.5\height}{\includegraphics[width=0.30\linewidth]{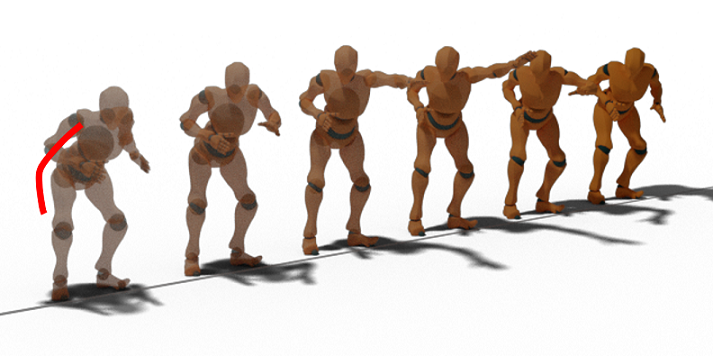}}}
\newcommand{\twocontent}{\raisebox{-.5\height}{\includegraphics[width=0.30\linewidth]{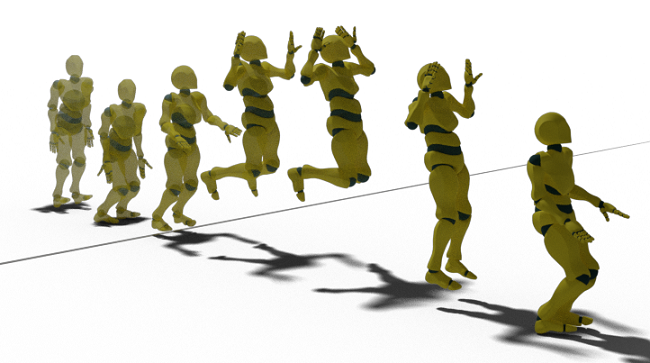}}}
\newcommand{\twoAberman}{\raisebox{-.5\height}{\includegraphics[width=0.30\linewidth]{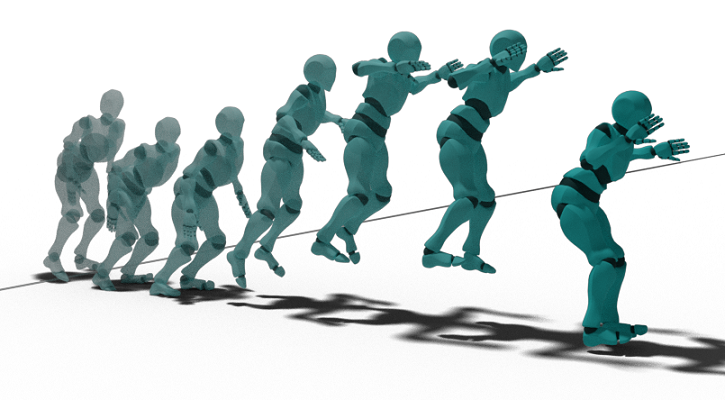}}}
\newcommand{\twoPark}{\raisebox{-.5\height}{\includegraphics[width=0.30\linewidth]{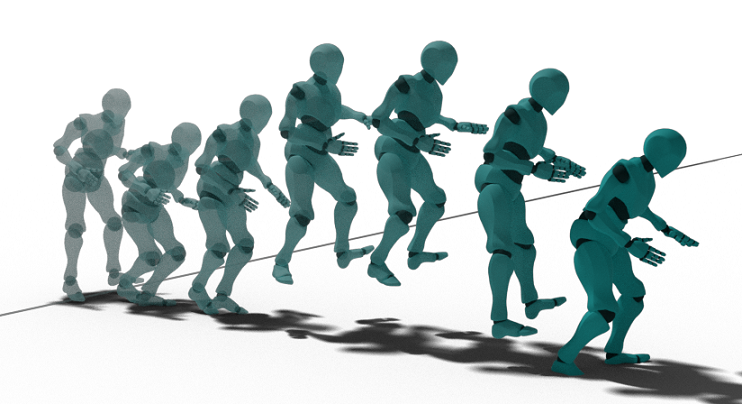}}}
\newcommand{\twoWen}{\raisebox{-.5\height}{\includegraphics[width=0.30\linewidth]{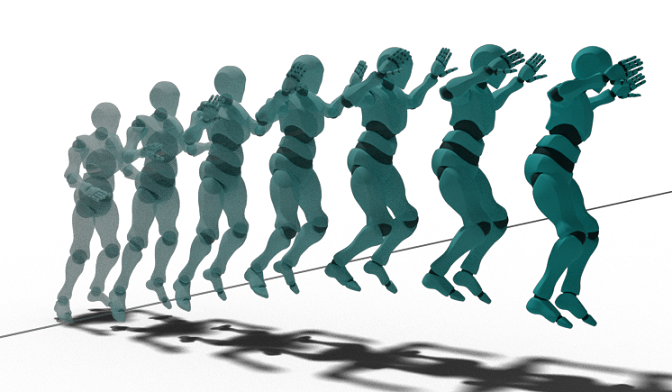}}}
\newcommand{\twoPuzzle}{\raisebox{-.5\height}{\includegraphics[width=0.30\linewidth]{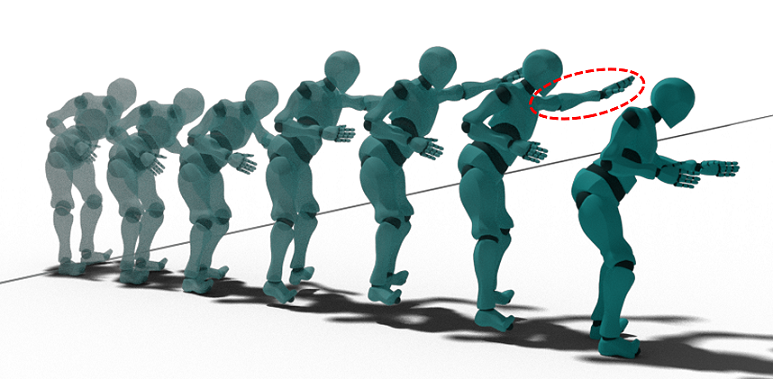}}}
\newcommand{\twoOURS}{\raisebox{-.5\height}{\includegraphics[width=0.30\linewidth]{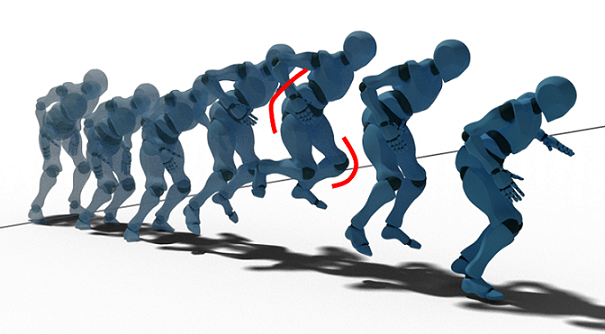}}}

\newcommand{\thrstyle}{\raisebox{-.5\height}{\includegraphics[width=0.30\linewidth]{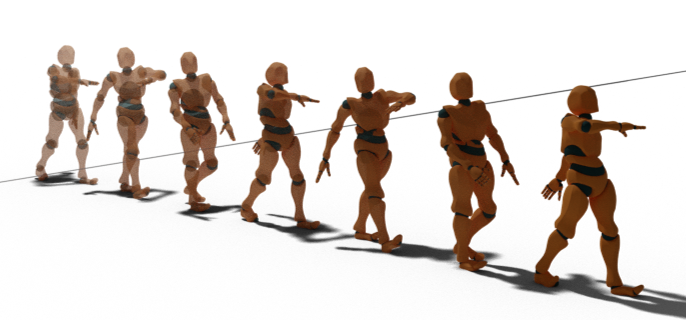}}}
\newcommand{\thrcontent}{\raisebox{-.5\height}{\includegraphics[width=0.30\linewidth]{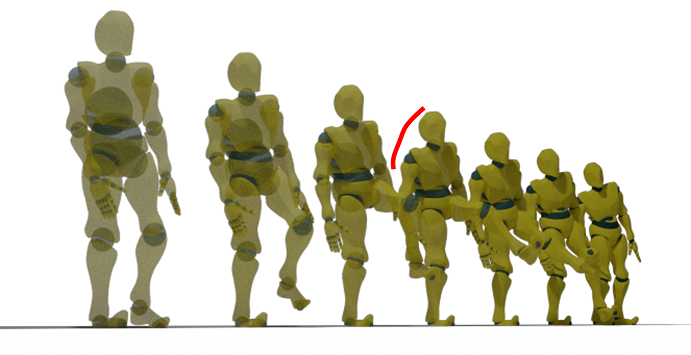}}}
\newcommand{\thrAberman}{\raisebox{-.5\height}{\includegraphics[width=0.30\linewidth]{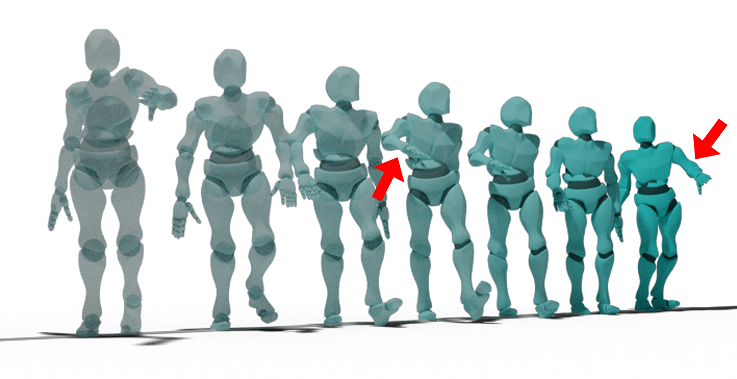}}}
\newcommand{\thrPark}{\raisebox{-.5\height}{\includegraphics[width=0.30\linewidth]{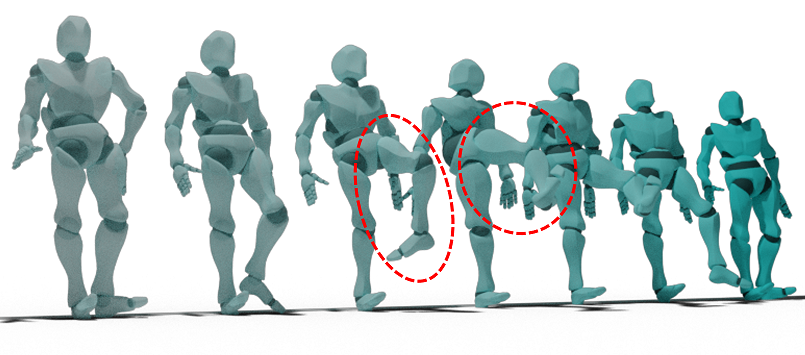}}}
\newcommand{\thrWen}{\raisebox{-.5\height}{\includegraphics[width=0.30\linewidth]{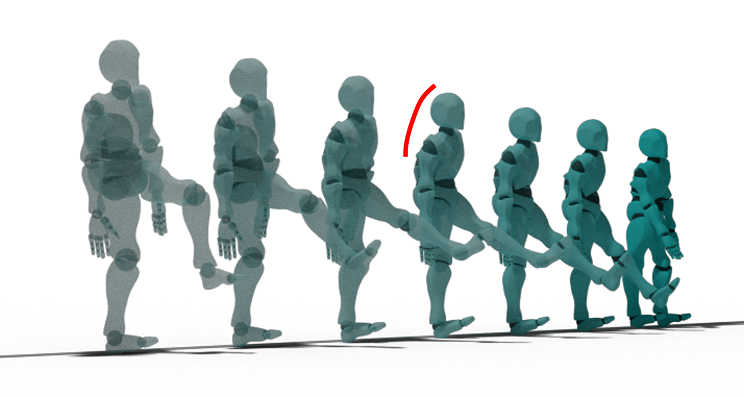}}}
\newcommand{\thrPuzzle}{\raisebox{-.5\height}{\includegraphics[width=0.30\linewidth]{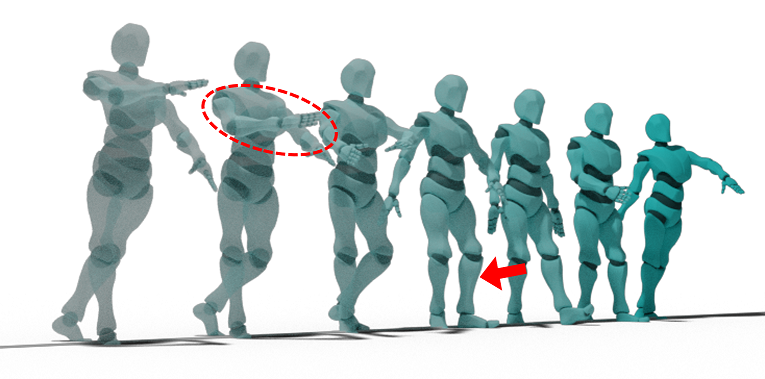}}}
\newcommand{\thrOURS}{\raisebox{-.5\height}{\includegraphics[width=0.30\linewidth]{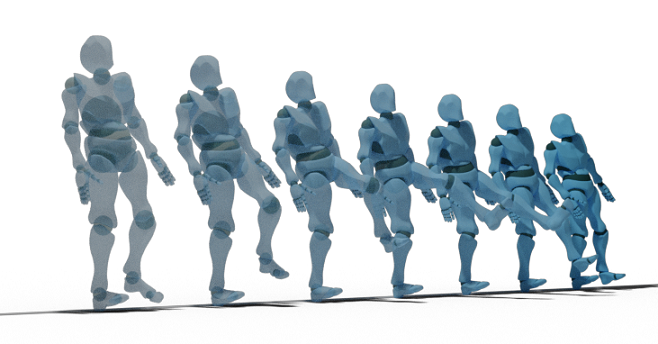}}}

\newcommand{\bfastyle}{\raisebox{-.5\height}{\includegraphics[width=0.30\linewidth]{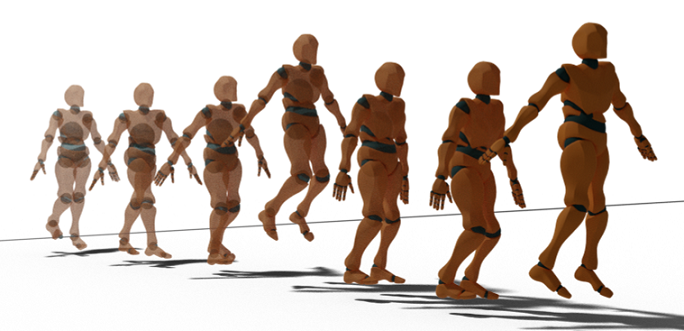}}}
\newcommand{\bfacontent}{\raisebox{-.5\height}{\includegraphics[width=0.30\linewidth]{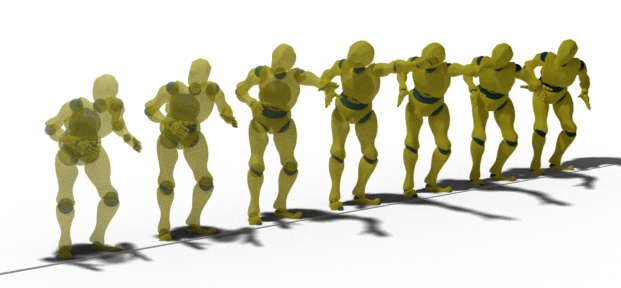}}}
\newcommand{\bfaAberman}{\raisebox{-.5\height}{\includegraphics[width=0.30\linewidth]{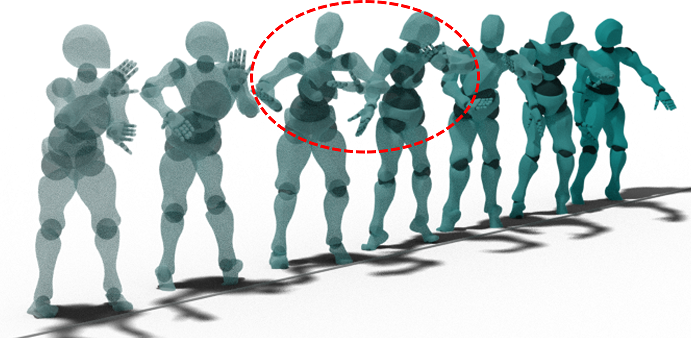}}}
\newcommand{\bfaPark}{\raisebox{-.5\height}{\includegraphics[width=0.30\linewidth]{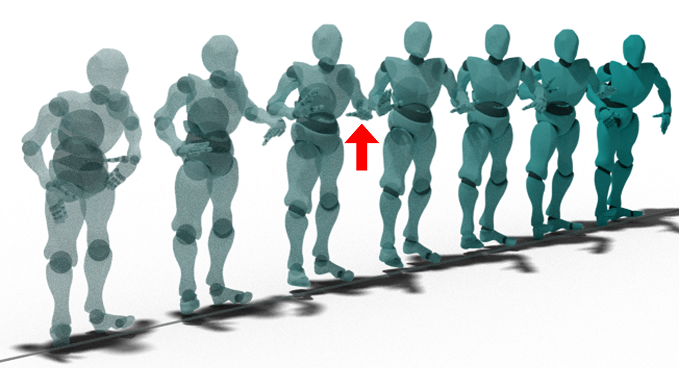}}}
\newcommand{\bfaWen}{\raisebox{-.5\height}{\includegraphics[width=0.30\linewidth]{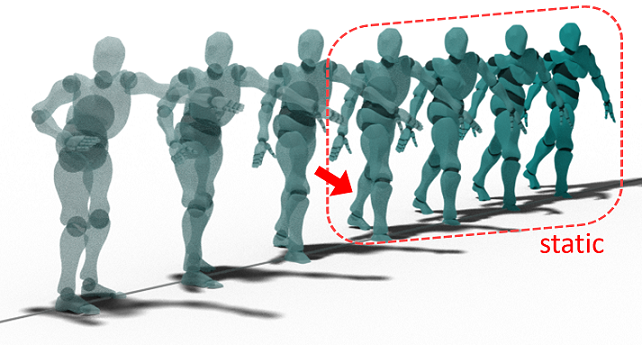}}}
\newcommand{\bfaPuzzle}{\raisebox{-.5\height}{\includegraphics[width=0.30\linewidth]{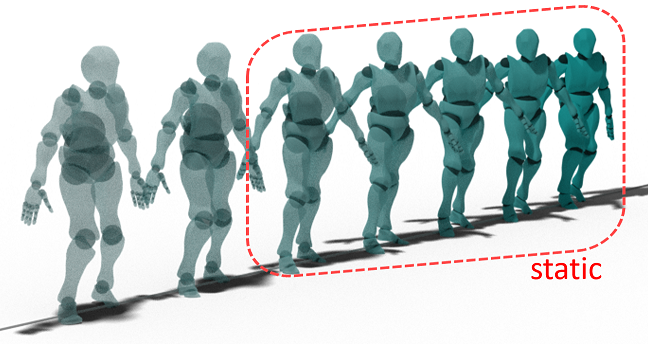}}}
\newcommand{\bfaOURS}{\raisebox{-.5\height}{\includegraphics[width=0.30\linewidth]{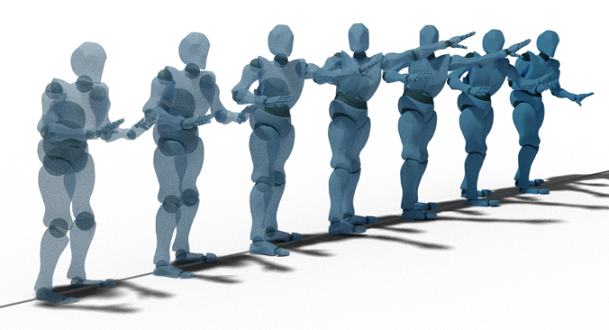}}}

\begin{figure*}[tb]
\centering
\resizebox{0.97\textwidth}{!}{%
    \begin{tabular}{c|cccc}
        & (1) & (2) & (3) & (4)  \\
        \midrule
        \begin{tabular}[c]{@{}c@{}}Input\\ style\\ motion\end{tabular} 
        & \begin{tabular}[c]{@{}c@{}}{\textbf{Old} Punch}\\\twostyle\end{tabular} 
        & \begin{tabular}[c]{@{}c@{}}{\textbf{Strutting} Kick}\\\onestyle\end{tabular} 
        & \begin{tabular}[c]{@{}c@{}}{\textbf{Proud} Walk}\\\thrstyle\end{tabular} 
        & \begin{tabular}[c]{@{}c@{}}{\textbf{FemaleModel} (BFA dataset)}\\\bfastyle \end{tabular} \\
        \begin{tabular}[c]{@{}c@{}}Input\\ content \\ motion\end{tabular} 
        & \begin{tabular}[c]{@{}c@{}}{Sexy \textbf{Jump}}\\\twocontent\end{tabular} 
        & \begin{tabular}[c]{@{}c@{}}{Childlike \textbf{Punch}}\\\onecontent\end{tabular} 
        & \begin{tabular}[c]{@{}c@{}}{Depressed \textbf{Kick}}\\\thrcontent\end{tabular} 
        & \begin{tabular}[c]{@{}c@{}}{Old \textbf{Punch}}\\\bfacontent\end{tabular} \\
        \midrule
        \textbf{MoST}  
        & \begin{tabular}[c]{@{}c@{}} {\textbf{Old Jump}} \\\twoOURS\end{tabular} 
        & \begin{tabular}[c]{@{}c@{}} {\textbf{Strutting Punch}} \\\oneOURS\end{tabular} 
        & \begin{tabular}[c]{@{}c@{}} {\textbf{Proud Kick}} \\\thrOURS\end{tabular} 
        & \begin{tabular}[c]{@{}c@{}} {\textbf{FemaleModel Punch}} \\\bfaOURS\end{tabular} 
        \\
        \midrule
        \begin{tabular}[c]{@{}c@{}}Aberman\\ \etal~\cite{aberman2020unpaired}\end{tabular} & \twoAberman & \oneAberman & \thrAberman & \bfaAberman \\
        \begin{tabular}[c]{@{}c@{}}Park\\ \etal~\cite{park2021diverse}\end{tabular}& \twoPark & \onePark & \thrPark & \bfaPark \\
        \begin{tabular}[c]{@{}c@{}}Wen\\ \etal~\cite{wen2021autoregressive}\end{tabular} & \twoWen & \oneWen & \thrWen & \bfaWen \\
        \begin{tabular}[c]{@{}c@{}}MotionPuzzle \\ ~\cite{jang2022motion}\end{tabular}& \twoPuzzle & \onePuzzle & \thrPuzzle & \bfaPuzzle \\
        \midrule
    \end{tabular}%
}
\caption{Qualitative results in Xia~\cite{xia2015realtime} and BFA~\cite{aberman2020unpaired} datasets. Please refer to the red indications. (1) Our method better reflects the style of \textit{old} in comparison to other existing methods, accurately representing both the bent upper body and leg. (2) Other methods fail to preserve the \cn{content} of \textit{punch}, instead, they result in peculiar leg movements or body twists. On the other hand, our result accurately depicts \textit{strutting punch}, where the upper body leans backward. (3) The results of \cite{aberman2020unpaired} and \cite{jang2022motion} do not exhibit a \textit{kick}, instead, their arm moves. \cite{park2021diverse} yields twisted leg movements. (4) Unlike our method, others fail to preserve the content of \textit{punch}, resulting in vibrations in static poses or twists
}
\label{fig:qualitative}
\end{figure*}
\subsection{Comparison with State-of-the-Art Methods}
We compare MoST to the state-of-the-art methods in motion style transfer, including the method of Aberman \etal~\cite{aberman2020unpaired}, Park \etal~\cite{park2021diverse}, Wen \etal ~\cite{wen2021autoregressive} and Jang \etal (MotionPuzzle)~\cite{jang2022motion}.
All the provided post-processing were employed in existing methods~\cite{aberman2020unpaired,park2021diverse,wen2021autoregressive, jang2022motion}, including copying global translation, removing foot skating, and correcting global velocity. 
In contrast, our result uses the raw output motion as-is.
Note that the method of Park \etal~\cite{park2021diverse} utilizes a style label in addition to the \textit{style motion} during both training and inference times, while other methods do not.

As indicated in Table~\ref{table:sota}, our method achieves the lowest values in both CC and SC\pp. Specifically, MoST exhibits a significant drop in CC.
This suggests that the well-disentangled \st{style} is injected into the appropriate body part without compromising the \cn{content} of $M^C$.
In particular, MoST demonstrates strong performance in both scenarios, where the input motions have the same \cn{content} or different \cn{contents}. In contrast, other methods exhibit a significant performance degradation when handling input motions with different \cn{contents}.
MotionPuzzle tends to generate motion by replicating $M^S$. The high CC value highlights this limitation, indicating that \cn{content} of $M^C$ is not preserved. For the same reason, MotionPuzzle shows a low SC\pp when the inputs have the same \cn{contents} but a significantly high SC\pp when they are different. 
The method of Aberman \etal~\cite{aberman2020unpaired} yields high CC and SC\pp scores, aligned with qualitative results that indicate style transfer failures in numerous cases.
The method of Park \etal~\cite{park2021diverse} yields a satisfactory result in SC\pp; however, there is a significant disparity in CC compared to our method. 
In their results, twisted motion is often observed in certain body parts.
Conversely, the method proposed by Wen \etal~\cite{wen2021autoregressive} demonstrates satisfactory results in CC but exhibits high SC\pp.
Their generated motions are usually smooth and realistic; however, they do not effectively reflect the style.

We present the qualitative results in Fig.~\ref{fig:qualitative}. 
We also have attached a video for better visualization in the supplementary material.
Despite not employing heuristic post-processing, our generated motions are both plausible and well-stylized. 
They exhibit clear body movements corresponding to the desired \cn{content}, while \st{style} is accurately expressed in the appropriate body parts.
In contrast, the methods of Aberman \etal, Park \etal, and MotionPuzzle tend to fail in preserving the \cn{content}, resulting in awkward motion where the \cn{contents} of the two inputs are blended in different body parts.
Moreover, it is difficult to discern the \st{style} expression in their results. 
While the method of Wen \etal produces plausible motions, it fails to effectively swap \st{styles}.
MoST exhibits favorable results for the unseen style during training, which is the style in BFA dataset~\cite{aberman2020unpaired}. However, other methods fail to preserve \cn{content} as shown in Fig.~\ref{fig:qualitative} (4)

\subsection{Self Study}
\begin{figure}[bt]
\centering
    \includegraphics[trim={1.7cm 1.7cm 1.7cm 1.7cm}, clip, width=0.9\linewidth]{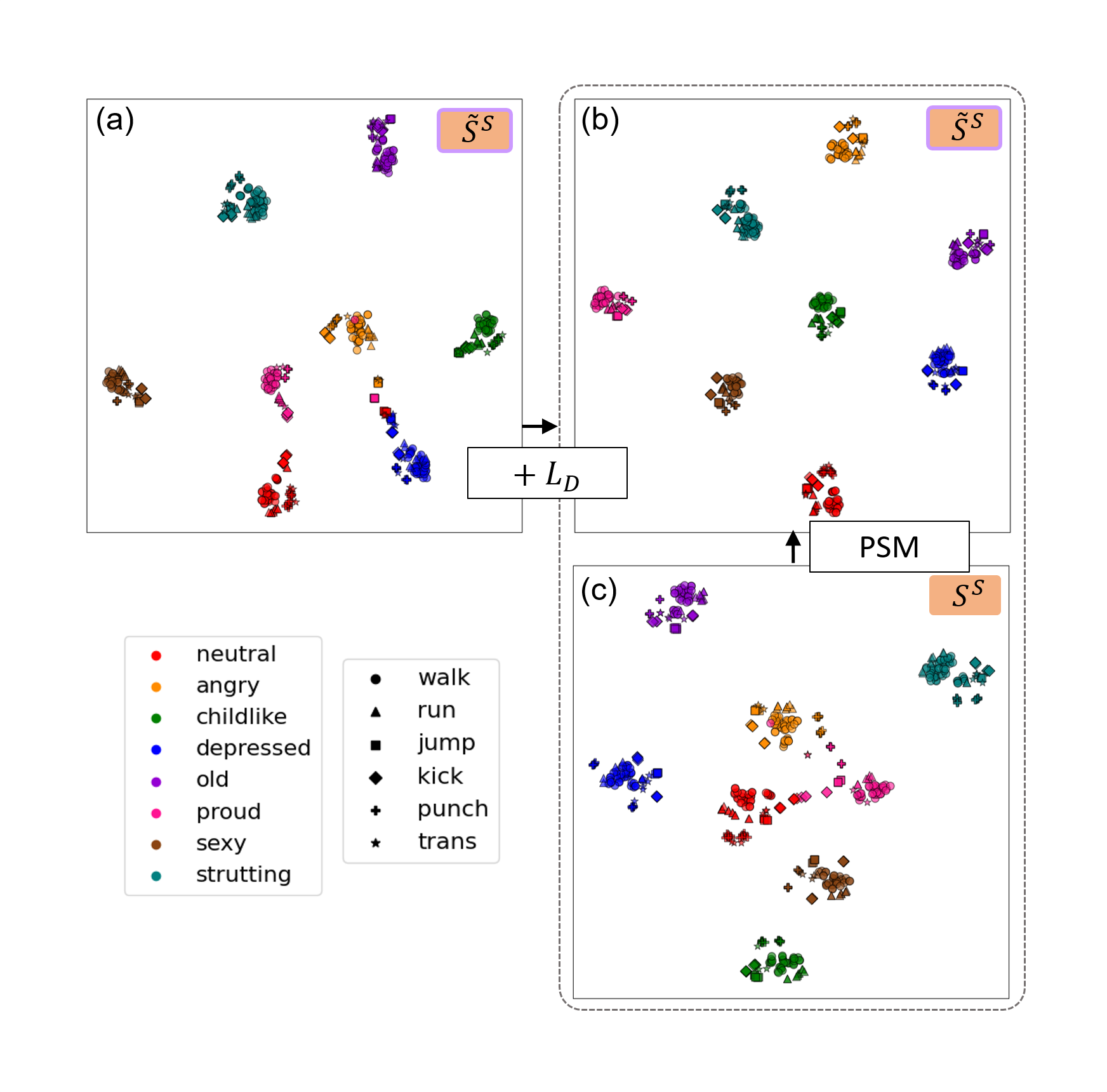}
    \caption{(a-b) Visualization of the modulated style feature ($\tilde{S}^S$) space of MoST in different loss settings. $L_{pre}$ and $L_{phy}$ are applied by default in (a). $L_{D}$ is additionally introduced in (b). All training and testing data are used as \textit{style motion}, and a single data point in the test set is employed for \textit{content motion}. The spaces are projected in 2D through t-SNE. The samples are visualized with different shapes according to their content labels and different colors according to their style labels. 
    (c) Space of $S^S$ before PSM. All loss functions are applied
    }
    \label{fig:abl_loss}
\end{figure}

\begin{figure}[bt]
\centering
    \includegraphics[trim={1.7cm 1.7cm 1.7cm 1.7cm}, clip, width=0.9\linewidth]{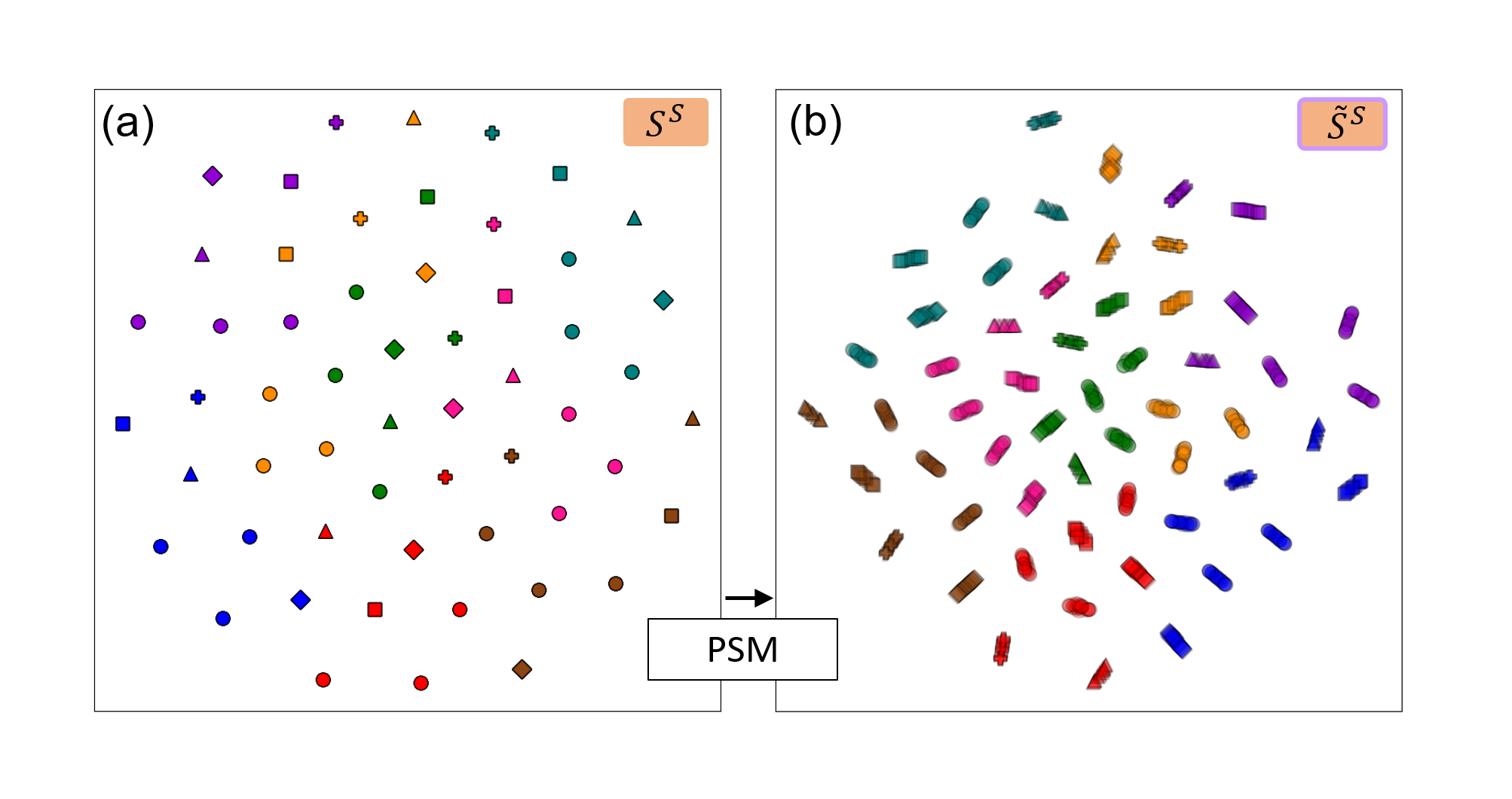}
    \caption{Visualization of $S^S$ and $\tilde{S}^S$ spaces projected through t-SNE for all 56 ($M^C$) $\times$ 56 ($M^S$) motion pairs in the test set. 
    }
    \label{fig:abl_psm}
\end{figure}

In Table~\ref{table:ablation}, we evaluate the impact of the proposed loss function and PSM. 
When PSM is applied, CC significantly decreases, confirming that the style modulation prevents the generation of movement in undesired body parts. $L_{D}$ substantially reduces both CC and SC\pp, highlighting the importance of disentangling \st{style} and \cn{content}.
The model achieves the best performance when both $L_{D}$ and PSM are employed.
\begin{table}[tb]
\caption{Ablation study for verifying the proposed $L_{D}$, and PSM on Xia dataset~\cite{xia2015realtime}. $L_{pre}$ and $L_{phy}$ are applied by default.
}
\label{table:ablation}
\centering
\resizebox{0.45\linewidth}{!}{%
    \begin{tabular}{cc|c|c}
        \toprule
          PSM & $L_{D}$ & CC $\downarrow$ & SC\pp $\downarrow$ \\\midrule
           &   & 37.4  &   69.5 \\ 
         \checkmark &  & 19.7  &  66.1  \\       
           & \checkmark   & 9.3  &  63.2  \\    
         \checkmark  & \checkmark  & \textbf{8.5}  & \textbf{63.0}   \\    

        \bottomrule
    \end{tabular}%
}
\end{table}
The effectiveness of $L_{D}$ is further demonstrated in the visualization of the style feature space.
Fig.~\ref{fig:abl_loss} (a) illustrates the space of $\tilde{S}^S$ before applying $L_{D}$, where the style features have been mixed across certain categories.
Fig.~\ref{fig:abl_loss} (b) shows that $L_{D}$ enforces complete separation among different styles.
Furthermore, the transition from Fig.~\ref{fig:abl_loss} (c) to (b) shows that the raw style features become distinctive through PSM.
Fig.~\ref{fig:abl_psm} shows how PSM modulates $S^S$ according to different $M^C$.
Fig.~\ref{fig:abl_psm} (a) illustrates the space of $S^S$, where each point represents an overlap of every 56 $M^C$ instances, as $S^S$ is yielded independently of $M^C$.
After PSM, the data points are spread according to the paired $M^C$ in the $\tilde{S}^S$ space.

\begin{figure}[bt]
\centering
    \includegraphics[trim={1.3cm 1.3cm 1.3cm 1.3cm}, clip, width=0.85\linewidth]{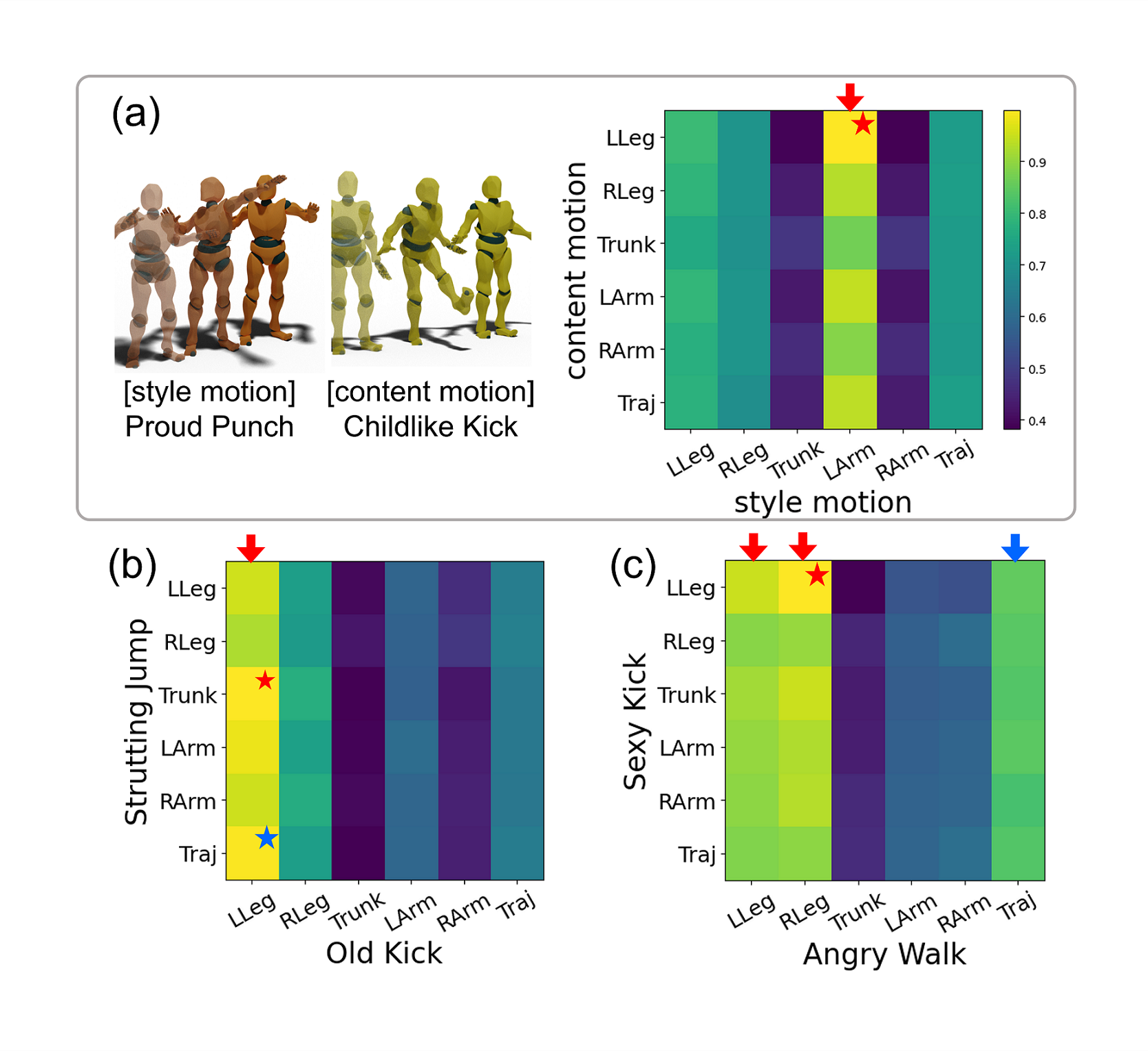}
    \caption{Cross-attention maps of PSM. The highest column (\includegraphics[height=1em]{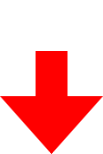}) indicates the body part in the \textit{style motion} from which the style originates. The highest element (\includegraphics[height=1em]{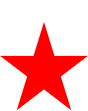}) pinpoints the body part in the \textit{content motion} that will receive the style. Traj. refers to global translation. The symbols \includegraphics[height=1em]{figures/indicators/blue_arrow.png} and \includegraphics[height=1em]{figures/indicators/blue_star.png} are indicators related to traj.
    }
    \label{fig:psm_viz}
\end{figure}

In Fig.~\ref{fig:psm_viz}, we have visualized cross-attention maps of PSM.
PSM identifies the activated body parts and transfers style from/to those parts.
Specifically, in (a), style of the arm in the `Punch' motion is transferred to the leg in the `Kick,' as indicated by the attention map.
In (b) and (c), high attention values are observed for the left leg in the 'Kick' and both legs in the 'Walk,' respectively.
Notably, PSM incorporates global translation alongside body parts.
The symbol \includegraphics[height=1em]{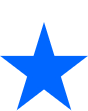} in (b) represents the application of style to traj, while \includegraphics[height=1em]{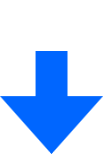} in (c) represents the origin of style from traj.

\section{Conclusions}
MoST is designed to effectively disentangle style and content of input motions and transfer styles between them.
The proposed loss functions successfully train MoST to generate well-stylized motion without compromising content.
Our method outperforms existing methods significantly, especially in motion pairs with different contents.
\\
\noindent
\textbf{Limitations and Future Works.} Some samples may exhibit the foot contact problem, as the physics-based loss does not forcibly eliminate it. Testing time optimization could be applied for the perfect removal.
While the transformer architecture offers advantages in handling sequences of diverse lengths, it requires specifying the maximum length of the model.
We plan to expand the model for few-shot learning, to handle small datasets due to the high cost of motion data acquisition.
\vspace{0.1cm}
\\
\textbf{Acknowledgement}
\\
This work was supported by IITP grant funded by MSIT [No.RS-2022-00164860] and [No.2021-0-01343].
\clearpage
{
    \small
    \bibliographystyle{ieeenat_fullname}
    \bibliography{main}
}
\clearpage

\clearpage
\setcounter{page}{1}
\maketitlesupplementary
\renewcommand\thesection{\Alph{section}}
\renewcommand\thesubsection{\thesection.\Alph{subsection}}
\renewcommand{\thefigure}{\Alph{figure}}
\renewcommand{\thetable}{\Alph{table}}
\setcounter{section}{0}
\setcounter{figure}{0}

\section{Additional Visualization of Limitations in Existing Methods}

Fig.\ref{fig:post_processing} illustrates the \textit{correction of foot contacts} which is employed in existing methods\cite{aberman2020unpaired, park2021diverse, jang2022motion} as a post-processing step. The raw outputs from these methods often suffer from issues such as floor penetration or foot skating, prompting the use of heuristic post-processing. In the \textit{correction of foot contacts} scheme, they extract foot contact information from the \textit{content motion}, use this data to adjust the positions of the feet, and apply inverse kinematics for output motion rectification.
In contrast, our method generates plausible motion without using any heuristic post-processing.

Fig.\ref{fig:limit_supple} illustrates representative failure cases of existing methods, complementing Fig.\ref{fig:limitations}. In Fig.\ref{fig:limit_supple} (a), we observe the corruption of the generated motion due to a lack of clear disentanglement between \st{style} and \cn{content}. The resulting motion fails to preserve the \cn{content} and struggles to express \st{style} effectively. Furthermore, in Fig.\ref{fig:limit_supple} (b), we examine the result from Wen \etal~\cite{wen2021autoregressive}, which encounters difficulties in expressing style. The method by Wen \etal~\cite{wen2021autoregressive} faces limitations, especially in handling short videos, attributed to its reliance on several front frames as a seed. As demonstrated in both Fig.~\ref{fig:limit_supple} (a) and (b), our method consistently produces well-stylized and plausible outputs.

\newcommand{\beforePark}{\raisebox{-.5\height}{\includegraphics[width=0.7\linewidth]{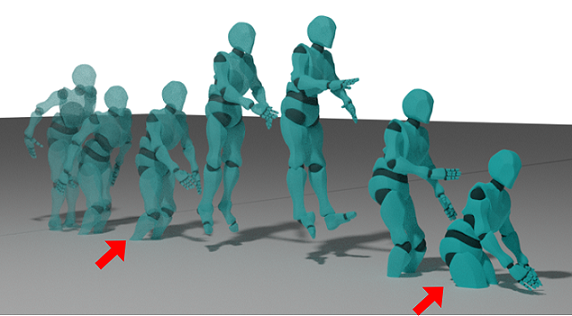}}}
\newcommand{\afterPark}{\raisebox{-.5\height}{\includegraphics[width=0.7\linewidth]{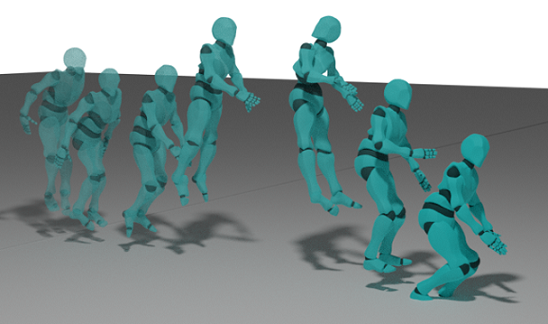}}}
\newcommand{\beforePuzzle}{\raisebox{-.5\height}{\includegraphics[width=0.7\linewidth]{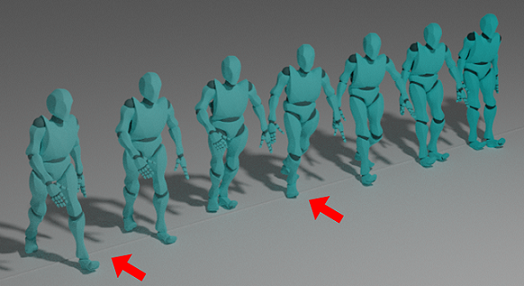}}}
\newcommand{\afterPuzzle}{\raisebox{-.5\height}{\includegraphics[width=0.7\linewidth]{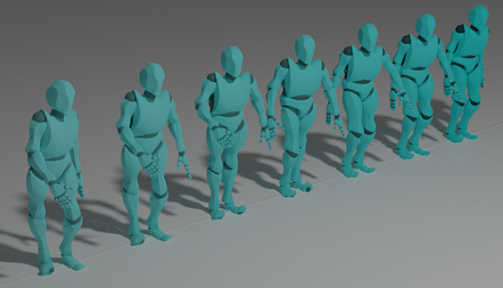}}}

\begin{figure}[ht]
\begin{minipage}{\linewidth}
    \centering
    \begin{tabular}{c}
        \begin{tabular}[c]{@{}c@{}}\beforePark\\{\small{Before post-processing}}\end{tabular} \\
        \begin{tabular}[c]{@{}c@{}}\afterPark\\{\small{After post-processing}}\end{tabular} 
    \end{tabular}%
    \vspace{-0.3cm}
    \caption*{(a)} 
\end{minipage}
\vspace{1.8cm}
\begin{minipage}{\linewidth}
    \centering
    \begin{tabular}{c}
        \begin{tabular}[c]{@{}c@{}}\beforePuzzle\\{\small{Before post-processing}}\end{tabular} \\
        \begin{tabular}[c]{@{}c@{}}\afterPuzzle\\{\small{After post-processing}}\end{tabular} 
    \end{tabular}%
    \vspace{-0.3cm}
    \caption*{(b)} 
\end{minipage}  
\vspace{-2.0cm}
\caption{Post-processing in existing methods. (a) The 'Depressed Jump' motions generated by Park \etal~\cite{park2021diverse}. In the output motion sequence of the model, the body penetrates the floor. Post-processing is applied to force correction and make contact with the ground. (b) The `Angry Punch' motion generated by MotionPuzzle~\cite{jang2022motion} network contains feet movement. 
Post-processing is employed to enforce the fixity of the feet
} 
\vspace{-0.3cm}
\label{fig:post_processing}
\end{figure}
\newcommand{\parkCnt}{\raisebox{-.5\height}{\includegraphics[width=0.30\linewidth]{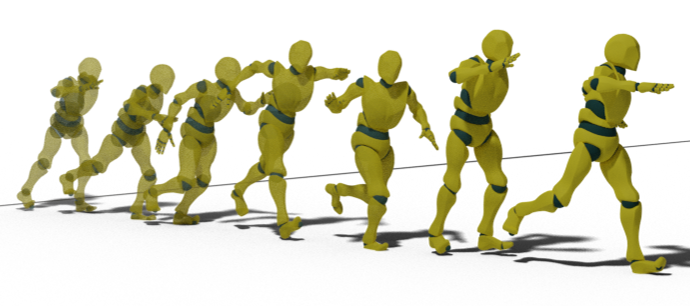}}}
\newcommand{\parkSty}{\raisebox{-.5\height}{\includegraphics[width=0.30\linewidth]{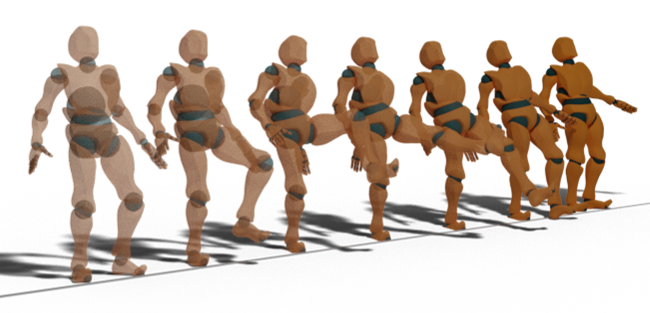}}}
\newcommand{\parkPark}{\raisebox{-.5\height}{\includegraphics[width=0.30\linewidth]{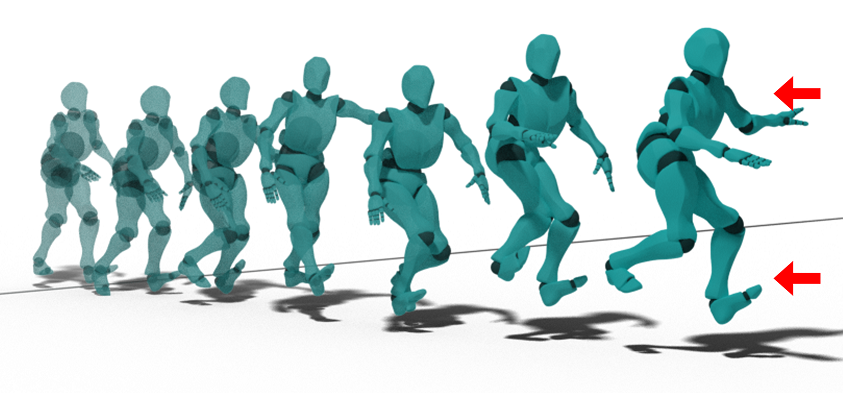}}}
\newcommand{\parkOurs}{\raisebox{-.5\height}{\includegraphics[width=0.30\linewidth]{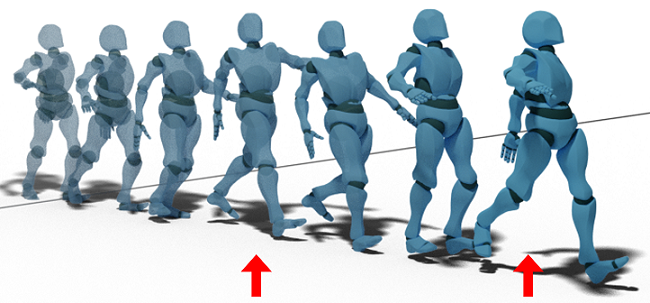}}}

\newcommand{\wenCnt}{\raisebox{-.5\height}{\includegraphics[width=0.30\linewidth]{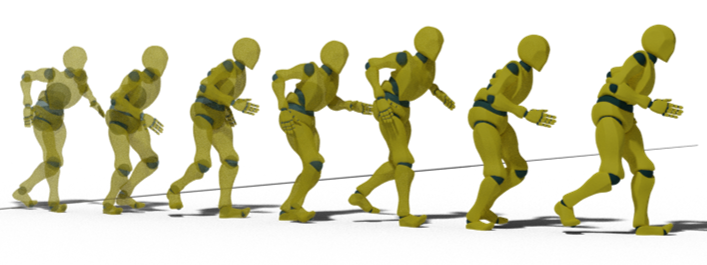}}}
\newcommand{\wenSty}{\raisebox{-.5\height}{\includegraphics[width=0.30\linewidth]{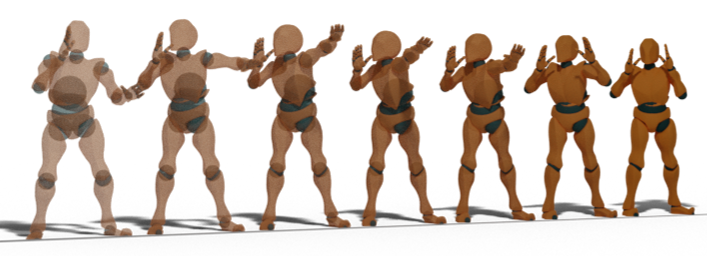}}}
\newcommand{\wenWen}{\raisebox{-.5\height}{\includegraphics[width=0.30\linewidth]{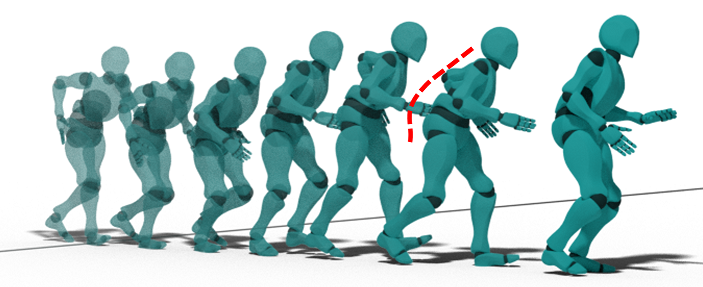}}}
\newcommand{\wenOurs}{\raisebox{-.5\height}{\includegraphics[width=0.30\linewidth]{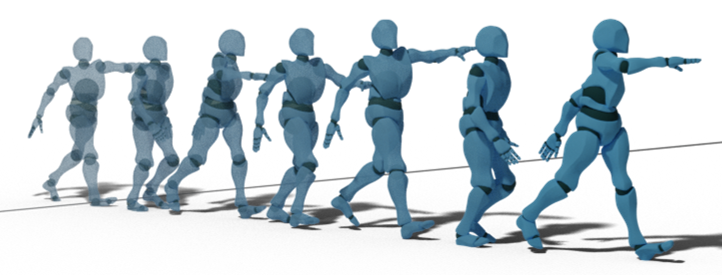}}}

\begin{figure*}[t]
\centering{
\begin{minipage}{0.92\linewidth}
    \centering
    \begin{tabular}{c|cc}
        \toprule
        \begin{tabular}{lr}  & \textit{Style Motion} \\&\\&\\&\\&\\ \textit{Content Motion} & \end{tabular} 
        & \multicolumn{2}{c}{\begin{tabular}[c]{@{}c@{}} {\textbf{Strutting} Kick} \\\parkSty \end{tabular} }
        \\\midrule
        \begin{tabular}[c]{@{}c@{}} {Angry \textbf{Run}} \\\parkCnt \end{tabular}
        & \begin{tabular}[c]{@{}c@{}} \parkPark\\{Park \etal~\cite{park2021diverse}}\end{tabular} 
        & \begin{tabular}[c]{@{}c@{}} \parkOurs\\{\textbf{MoST}} \end{tabular} 
        \\\bottomrule
    \end{tabular}%
    \vspace{-0.3cm}
    \caption*{(a)}
\end{minipage}
\vspace{0.2cm}

\begin{minipage}{0.92\linewidth}
    \centering
    \begin{tabular}{c|cc}
        \toprule
        \begin{tabular}{lr}  & \textit{Style Motion} \\&\\&\\&\\&\\ \textit{Content Motion} & \end{tabular} 
        & \multicolumn{2}{c}{\begin{tabular}[c]{@{}c@{}} {\textbf{Proud} Punch} \\\wenSty \end{tabular} }
        \\\midrule
        \begin{tabular}[c]{@{}c@{}} {Old \textbf{Run}} \\\wenCnt \end{tabular}
        & \begin{tabular}[c]{@{}c@{}} \wenWen\\{Wen \etal~\cite{wen2021autoregressive}}\end{tabular} 
        & \begin{tabular}[c]{@{}c@{}} \wenOurs\\{\textbf{MoST}} \end{tabular} 
        \\\bottomrule
    \end{tabular}%
    \vspace{-0.3cm}
    \caption*{(b)}
\end{minipage}
\vspace{-0.3cm}
}
\caption{Representative failure cases in existing methods: (a) A result from Park \etal~\cite{park2021diverse} shows corrupted motion, with the generated animation depicting the leg swinging in mid-air and the arms losing movement. (b) A result from Wen \etal~\cite{wen2021autoregressive} exhibits issues in expressing style. For a clearer visualization, please refer to the attached video.}
\label{fig:limit_supple}

\end{figure*}

\section{Details of Motion Representation}
In this section, we explain more details about the motion representation in Sec.~\ref{sec:motion_representation}.
$m_t^j$ represents $j$-th joint vector at $t$-th frame.
The joint vector is represented by $m_t^j = [o_t^j; q_t^j]$, where $o_t^j \in \mathbb{R}^3$ indicates the 3-dimensional vector of each joint from the root joint and $q_t^j \in \mathbb{R}^4$ indicates joint rotation expressed by a unit quaternion. $q_t^j$ is based on the world axis set with the anterior direction of the body.
The global joint vector of $t$-th frame is represented as $m_t^{root} = [o_t^{root}; q_t^{root}]$, where $o_t^{root}$ and $q_t^{root}$ indicate the position of the root joint and the global rotation, respectively.
A global velocity vector $v_t = [\acute{o}_t^{root}; a_t]$ is additionally used for the global motion following~\cite{park2021diverse} and~\cite{wen2021autoregressive}, where $\acute{o}_t^{root} \in \mathbb{R}^3$ and $a_t^{root} \in \mathbb{R}$ denotes a positional velocity and an angular velocity of the root joint, respectively.

\section{Details of Adopted Loss}
We provide details of adopted losses from existing methods in this section.
Adversarial loss~\cite{aberman2020unpaired,park2021diverse} is written as
\begin{align}
    L_{adv} &= \mathbb{E}_{M^S \sim \mathbb{M}} \; [ log(\mathcal{D}(M^S)) ]\\
    &+ \mathbb{E}_{M^C, M^S \sim \mathbb{M}} \; [log(1-\mathcal{D}(M^G))]. \notag
\end{align}
Reconstruction loss~\cite{aberman2020unpaired,park2021diverse} is written as
\begin{align}
    L_{recon} &= \mathbb{E}_{M^C\sim \mathbb{M}} \; || \mathtt{MoST}(M^C, M^C) - M^C ||_2.
\end{align}
Cycle consistency loss~\cite{jang2022motion} is written as
\begin{align}
    L_{cyc} &= L_{cyc\text{-}s} + L_{cyc\text{-}c}, \\
    L_{cyc\text{-}s} &= \mathbb{E}_{M^C, M^S \sim \mathbb{M}} \; ||\;\mathtt{MoST}(M^S, M^G) - M^S ||_2, \\
    L_{cyc\text{-}c} &= \mathbb{E}_{M^C, M^S \sim \mathbb{M}} \; ||\;\mathtt{MoST}(M^G, M^C) - M^C ||_2, \\
    M^G &= \mathtt{MoST}(M^C, M^S),
\end{align}

Inspired by~\cite{rempe2020contact}, we introduce velocity and acceleration regularization in Eq.(\ref{eq:phy}) for the generated motion $M^G$ as
\begin{align}
    R_{vel} &= \frac{1}{T-1}\sum_{t=1}^{T-1} \left( \sum_{j=1}^{J} ||\dot{\hat{m}}_{t}^j||_2 + ||\dot{\hat{m}}_{t}^{root}||_2 \right), \\
    R_{acc} &= \frac{1}{T-2}\sum_{t=1}^{T-2} \left( \sum_{j=1}^{J} ||\ddot{\hat{m}}_{t}^j||_2 + ||\hat{v}_{t+1}-\hat{v}_{t}||_2 \right), \notag
\end{align}
where $\dot{\hat{m}}_{t}^j = \hat{m}_{t+1}^j - \hat{m}_{t}^j$, $\dot{\hat{m}}_{t}^{root} = \hat{m}_{t+1}^{root} - \hat{m}_{t}^{root}$ and $\ddot{\hat{m}}_{t}^j  = \dot{\hat{m}}_{t+1}^j  - \dot{\hat{m}}_{t}^j $.

\begin{figure*}[ht]
\centering
    \includegraphics[width=0.75\linewidth]{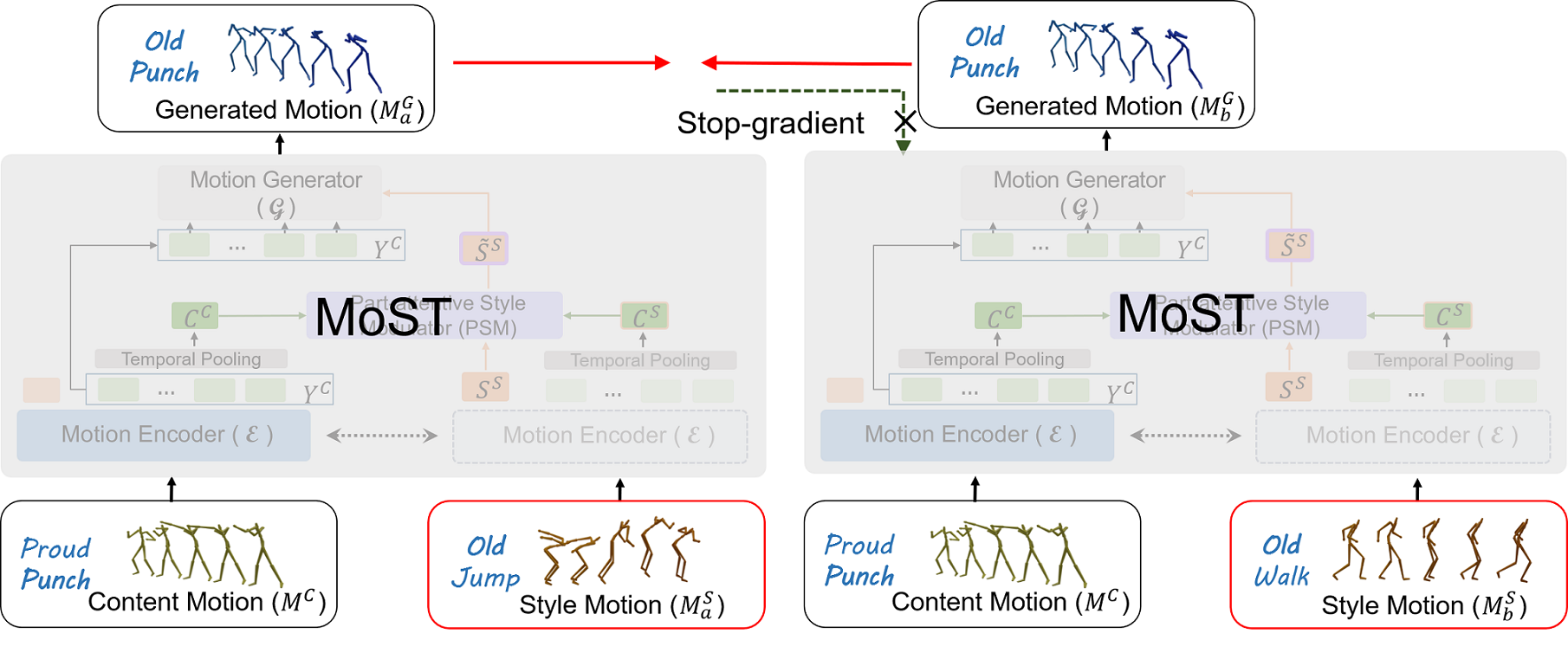}
    \caption{Illustration explaining the proposed style disentanglement loss ($L_{D}$)}
    \label{fig:loss_disentangle}
\end{figure*}

\section{Details of Proposed Loss}
\label{sec:proposed_loss}
Fig.~\ref{fig:loss_disentangle} illustrates the proposed \textit{style disentanglement loss} ($L_{D}$).
We generate motions from two different \textit{style motions} with identical style labels but different content labels. $L_{D}$ makes both generated motions similar. For more stable training, we halt back propagation for the path to $\mathtt{MoST}(M^C, M^S_{b})$.


\begin{table*}[t!]
\caption{Ablation study of replacing $\mathcal{E}$, $\mathcal{G}$, and $\mathcal{D}$ with the different network architectures proposed by~\cite{aberman2020unpaired} and~\cite{park2021diverse}. Applied components are labeled as $\circ$. Note that the original method by Park \etal~\cite{park2021diverse} utilizes a style label, whereas our method does not.}
\label{table:ablation_net}
\centering
\resizebox{0.8\linewidth}{!}{%
    \begin{tabular}{l|l|cccc|c|c}
        \toprule
        & Networks of $\mathcal{E}$, $\mathcal{G}$, $\mathcal{D}$ & Style Label & Siamese Encoder & PSM & $\mathbf{L_{D}}$ & CC $\downarrow$  & SC\pp $\downarrow$\\ \midrule
        Aberman \etal~\cite{aberman2020unpaired}  & Temporal-CNN  & $\times$ & $\times$ & $\times$  & $\times$ & 46.0 & 189.7 \\
        OURS  & Temporal-CNN                                      & $\times$ & $\times$ & $\times$  & \Large{$\circ$} & \textbf{39.6} & \textbf{81.2} \\
        \midrule
        Park \etal~\cite{park2021diverse} &   GCN                 & \Large{$\circ$} & $\times$ & $\times$ & $\times$ & 38.4 & \textbf{65.7}  \\
        OURS & GCN                                                & $\times$ & $\times$ & \Large{$\circ$}& \Large{$\circ$} & \textbf{25.1} & 68.5 \\
        \midrule
        OURS (MoST) & Transformer                                 & $\times$ & \Large{$\circ$} & \Large{$\circ$} & \Large{$\circ$} & \textbf{8.5}  & \textbf{63.0} \\ 
        \bottomrule
    \end{tabular}%
}
\end{table*}

\section{Details of Evaluation Metrics}
In this section, we provide equations of evaluation metrics in Sec.~\ref{sec:experiments}.
We evaluate $M^G$ which is generated from a motion pair of $M^C$ and $M^S$ drawn from the test dataset.
$l_{C^C}$ and $l_{S^C}$ denote the content label and style label of $M^C$, respectively. $l_{C^S}$ and $l_{S^S}$ denote the content label and style label of $M^S$, respectively.
$l_{C^T}$ and $l_{S^T}$ denote the content label and style label of a ground truth motion sequence $M^T$ in a training sample, respectively.
Metrics are written as
\begin{align}
    \text{CC} &=  \mathbb{E}_{M^C, M^S \sim \{m_{test} | l_{S^C} = l_{S^S}\}} || M^G - M^C ||_2, \\
    \text{SC} &=  \mathbb{E}_{M^C, M^S \sim \{m_{test} | l_{C^C} = l_{C^S}\}} || M^G - M^S ||_2, \\
\begin{split}
    \text{SC\pp} &=  \mathbb{E}_{M^C, M^S \sim \{m_{test}\}} \\
    &\left(\mathbb{E}_{M^T \sim \{m|l_{C^T}=l_{C^C},~ l_{S^T}=l_{S^S}\}}|| M^G - M^T ||_2 \right), \notag
\end{split}
\end{align}
where $m_{test}$ denotes a random variable of motion data in the test set.
$m$ denotes a random variable of motion data in the training set.
The global translation is excluded when calculating the metric for a fair comparison. 

\section{Ablation Study in Architectures}
In Table~\ref{table:ablation_net}, we conduct an experiment by replacing the proposed transformer architecture in the encoder, generator, and discriminator with other architectures. Specifically, we employ temporal-CNN and GCN architectures, sourced from the methods of Aberman \etal~\cite{aberman2020unpaired} and Park \etal~\cite{park2021diverse}, respectively.
In both architectures, our proposed components couldn't be entirely implemented due to their structural limitations. 
The temporal-CNN~\cite{aberman2020unpaired} and GCN~\cite{park2021diverse} architectures lack the capability to extract both style and content features within a single network. 
Consequently, we constructed separate content encoders ($\mathcal{C}$) and style encoders ($\mathcal{S}$).
Additionally, obtaining part-specific features from the temporal-CNN~\cite{aberman2020unpaired} was not feasible. As a result, PSM couldn't be utilized in this case. Nevertheless, both new losses were applied.

Each network benefited from our overall framework and loss functions, surpassing the individual performances of the original methods listed in Table~\ref{table:sota}. Notably, the temporal-CNN achieved significantly lower values in both CC and SC\pp. Importantly, our proposed transformer architecture demonstrated superior performance compared to the other two networks.

\section{Evaluation Across Motion Categories}
\begin{table*}[ht]
\caption{Content consistency (CC) in each content category and in each style category on Xia dataset~\cite{xia2015realtime}. }
\label{table:CC_cat}
\centering

\resizebox{\linewidth}{!}{%
{\huge
    \begin{tabular}{l|k|k|k|k|k?c|c|c|c|c|c|c|c?c}
    \toprule
    \multirow{2}{*}{Methods}& \multicolumn{5}{c?}{Content categories of content motion} & \multicolumn{9}{c}{Style categories of style motion} \\ 
    \cmidrule(lr){2-6} \cmidrule(lr){7-15}
     & walk  & run & jump & kick & punch & neutral & angry & childlike & depressed &  old & proud & sexy & strutting & \textbf{average}\\ \midrule
    MotionPuzzle~\cite{jang2022motion}         & 37.0 & 50.5 & 68.8 & 63.7 & 65.9  & 51.9 & 57.9 & 56.4 & 49.9 & 42.4 & 59.3 & 52.0 & 41.8 & 51.4 \\
    Aberman \etal~\cite{aberman2020unpaired}   & 29.3 & 38.5 & 40.3 & 66.2 & 55.7  & 42.2 & 51.9 & 59.3 & 34.4 & 34.1 & 67.2 & 39.8 & 39.2 & 46.0 \\
    Park \etal~\cite{park2021diverse}          & 34.1  & 46.2 & 44.0 & 41.4 & 35.0  & 35.7 & 41.2 & 46.5 & 32.4 & 33.9 & 50.7 & 35.8 & 31.2 & 38.4 \\
    Wen \etal~\cite{wen2021autoregressive}     & 14.0 & 17.6 & 31.1 & 20.0 & 18.8 & 18.5 & \textbf{9.6} & 19.9 & 19.4 & 17.6 & 23.5 & 21.7 & 16.7 & 18.5 \\\midrule
    MoST  & \textbf{8.1} & \textbf{9.6} & \textbf{8.2} & \textbf{9.7} & \textbf{7.9} & \textbf{6.9} & 9.9 & \textbf{7.5} & \textbf{7.5} & \textbf{10.1} & \textbf{10.4} & \textbf{7.6} & \textbf{8.5} & \textbf{8.5}                                    \\
    \bottomrule
    \end{tabular}%
}
}
\vspace{0.3cm}
\caption{Style consistency\pp (SC\pp) in each content category and in each style category on Xia dataset~\cite{xia2015realtime}. }
\label{table:SC_cat}
\centering
\resizebox{\linewidth}{!}{%
{\huge
    \begin{tabular}{l|k|k|k|k|k?c|c|c|c|c|c|c|c?c}
    \toprule
    \multirow{2}{*}{Methods}& \multicolumn{5}{c?}{Content categories of content motion} & \multicolumn{9}{c}{Style categories of style motion} \\ 
    \cmidrule(lr){2-6} \cmidrule(lr){7-15}
     & walk  & run & jump & kick & punch & neutral & angry & childlike & depressed &  old & proud & sexy & strutting & \textbf{average} \\ \midrule
    MotionPuzzle~\cite{jang2022motion}         & 69.6 & 78.2 & 89.4 & 77.9 & 77.5 & 71.5 & 84.2 & 79.3 & 72.4 & 65.2 & 92.3 & 74.8 & 68.1 & 76.0 \\
    Aberman \etal~\cite{aberman2020unpaired}   & 258.9 & 237.1 & 115.0 & 112.8 & 85.9 & 185.1 & 174.0 & 209.9 & 181.3 & 182.2 & 196.9 & 195.3 & 192.5 & 189.7 \\
    Park \etal~\cite{park2021diverse}          & 65.2 & 77.9 & 64.8 & 68.3 & 53.3 & 59.2 & \textbf{66.5} & 70.5 & 59.8 & 62.0 & 81.1 & \textbf{62.4} & 64.2 & 65.7 \\
    Wen \etal~\cite{wen2021autoregressive}     & 72.7 & 87.5 & 97.4 & 97.5 & 65.0 & 71.1 & 79.5 & 84.5 & 77.2 & 83.8 & 87.2 & 78.6 & 84.5 & 80.8 \\\midrule
    MoST & \textbf{61.6} & \textbf{75.4} & \textbf{60.5} & \textbf{68.1} & \textbf{51.9} & \textbf{55.1} & 67.5 & \textbf{63.8} & \textbf{58.8} & \textbf{59.9} & \textbf{75.5} & 62.8 & \textbf{60.2} & \textbf{63.0} \\
    \bottomrule
    \end{tabular}%
}
}
\end{table*}
Table~\ref{table:CC_cat} and Table~\ref{table:SC_cat} present the CC and SC\pp in each content and style category.
Regarding \cn{content}, 'Run' and 'Kick' are proven to be challenging to retain.
For \st{style}, 'Proud,' 'Angry,' and 'Childlike' are proven to be difficult to express.

\section{Additional Results in BFA dataset}
Fig.~\ref{fig:qual_bfa} illustrates the additional results on BFA dataset~\cite{aberman2020unpaired}.
Our method effectively performs style transfer even when complex motions are mixed within a motion clip.
The figure demonstrates a clear differentiation of \st{styles} in the generated motions when different \st{styles} are transferred.
In addition, our method achieves robust transfer between two inputs with different \cn{contents}.

\section{Generation of Global Translation}
\begin{figure}[ht]
\centering
    \includegraphics[width=0.9\linewidth]{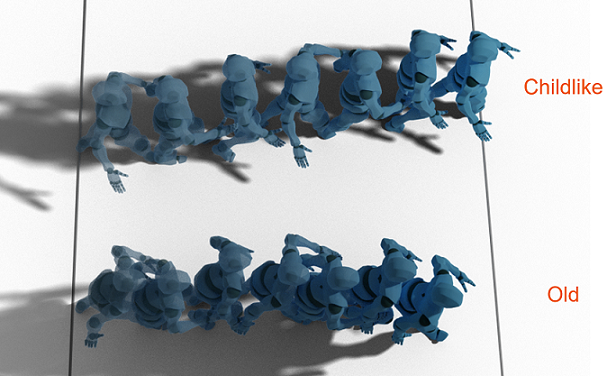}
    \caption{Generated global translation reflecting styles. (Top) `Childlike Walk' generated from `Neutral Walk' ($M^C$) and `Childlike Kick' ($M^S$). (Bottom) `Old Walk' generated from `Neutral Walk' ($M^C$) and `Old Kick' ($M^S$).}
    \label{fig:velocity}
\vspace{-0.3cm}
\end{figure}

Fig.\ref{fig:velocity} displays the global translation generated by our method. The figure illustrates the variation in motion speeds for different \st{styles} produced by our method. The upper motion is 'Childlike Walk,' while the lower motion is 'Old Walk.' The global translation of 'Old Walk' is generated at a slower speed. It is worth noting that Aberman \etal\cite{aberman2020unpaired} utilized a heuristic post-processing technique called \textit{global velocity warping} to diversify motion speeds between styles.

\newcommand{\failCnt}{\raisebox{-.5\height}{\includegraphics[width=0.46\linewidth]{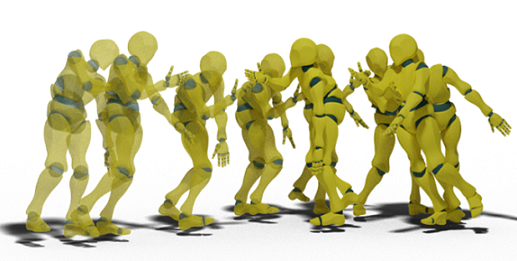}}}
\newcommand{\failOurs}{\raisebox{-.5\height}{\includegraphics[width=0.46\linewidth]{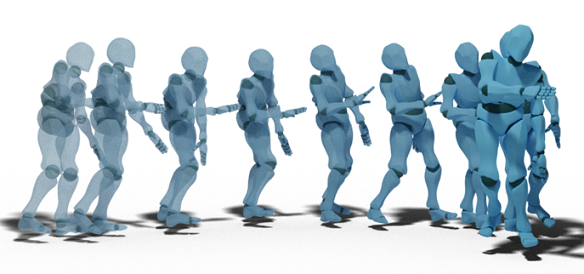}}}
\newcommand{\failSty}{\raisebox{-.5\height}{\includegraphics[width=0.46\linewidth]{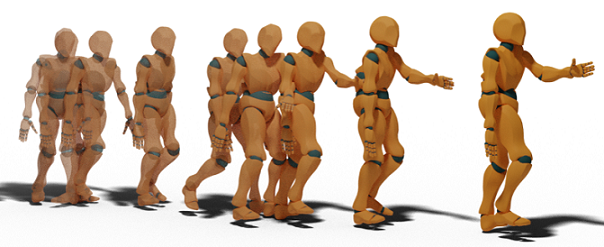}}}

\begin{figure}[ht]

\begin{minipage}{\linewidth}
    \centering
    \begin{tabular}{c|c}
        \toprule
        \begin{tabular}[c]{@{}c@{}} \textit{style} \\ \textit{motion} \end{tabular} &
        \begin{tabular}[c]{@{}c@{}} {\small{Angry}} \\\failSty \end{tabular} 
        \\\midrule
        \begin{tabular}[c]{@{}c@{}} \textit{content} \\ \textit{motion} \end{tabular} &
        \begin{tabular}[c]{@{}c@{}} {\small{Drunk}} \\\failCnt \end{tabular}
        \\\midrule
        \begin{tabular}[c]{@{}c@{}} \textit{generated} \\ \textit{motion} \end{tabular} &
        \begin{tabular}[c]{@{}c@{}} \failOurs \end{tabular}
        \\\bottomrule
    \end{tabular}%
    \vspace{-0.3cm}
    \caption*{(a)}
\end{minipage}
\\
\begin{minipage}{\linewidth}
    \centering
    \includegraphics[width=0.7\linewidth]{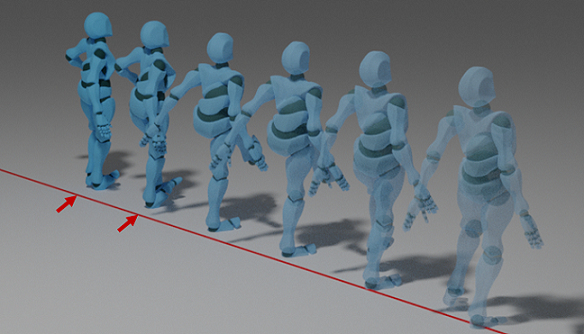}
    \vspace{-0.3cm}
    \caption*{(b)}
\end{minipage} 
\vspace{-0.3cm}
\caption{Failure cases obtained in our method. (a) Less successful style transfer for the intricate \textit{content motion}. (b) Foot skating observed in the generated motion of `Proud Kick.'} 
\vspace{-0.3cm}
\label{fig:failure_cases}
\end{figure}

\section{Failure Cases of MoST}
In Fig.\ref{fig:failure_cases}, we present the failure cases observed in our results. As shown in Fig.\ref{fig:failure_cases} (a), style transfer was relatively less successful for the \textit{content motion} featuring intricate motion, such as `Drunk.' In the generated motion, while the staggering appearance decreased, `Angry' was not expressed perfectly. Fig.~\ref{fig:failure_cases} (b) illustrates foot skating observed in the generated motion.

\newcommand{\fCnt}{\raisebox{-.5\height}{\includegraphics[width=0.30\linewidth]{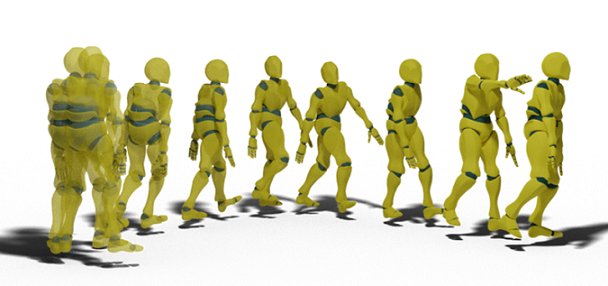}}}
\newcommand{\ffOurs}{\raisebox{-.5\height}{\includegraphics[width=0.30\linewidth]{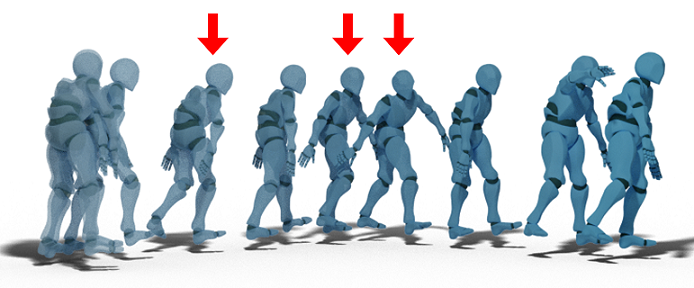}}}
\newcommand{\fsOurs}{\raisebox{-.5\height}{\includegraphics[width=0.30\linewidth]{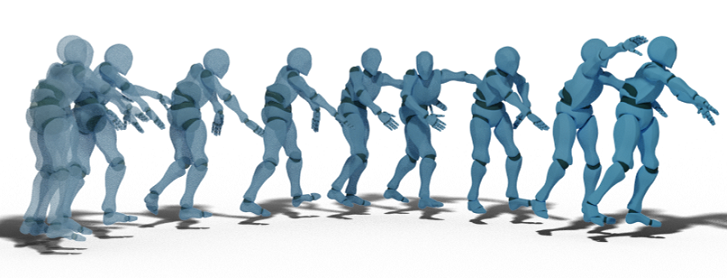}}}
\newcommand{\ffSty}{\raisebox{-.5\height}{\includegraphics[width=0.30\linewidth]{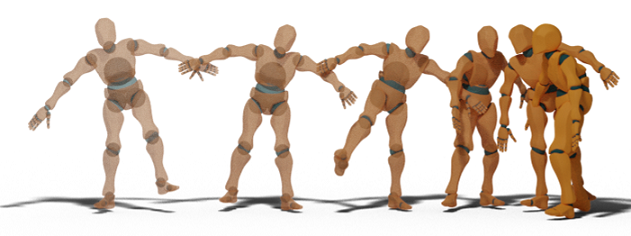}}}
\newcommand{\fsSty}{\raisebox{-.5\height}{\includegraphics[width=0.30\linewidth]{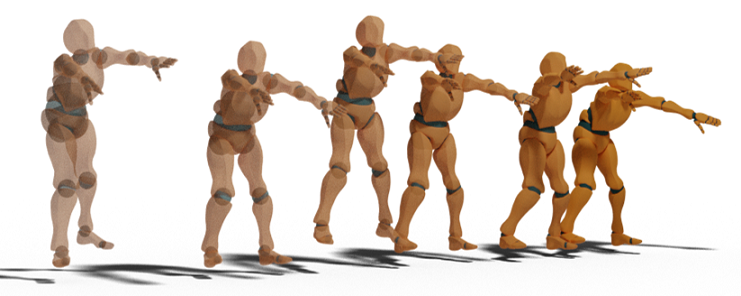}}}

\newcommand{\sCnt}{\raisebox{-.5\height}{\includegraphics[width=0.30\linewidth]{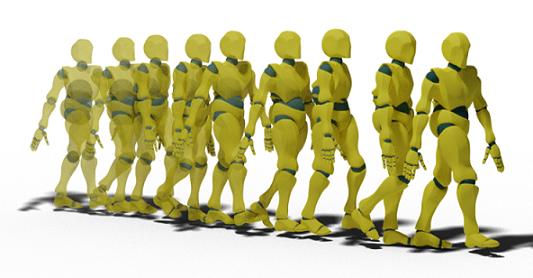}}}
\newcommand{\sfOurs}{\raisebox{-.5\height}{\includegraphics[width=0.30\linewidth]{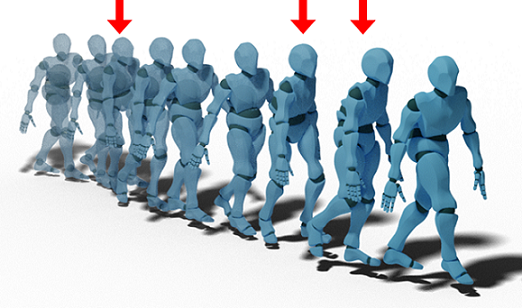}}}
\newcommand{\ssOurs}{\raisebox{-.5\height}{\includegraphics[width=0.30\linewidth]{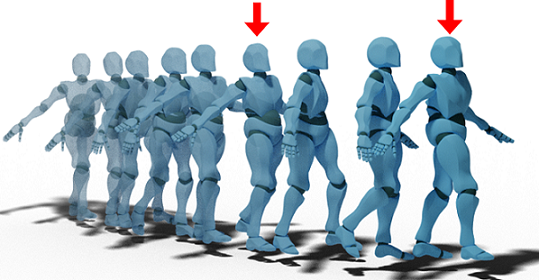}}}
\newcommand{\sfSty}{\raisebox{-.5\height}{\includegraphics[width=0.30\linewidth]{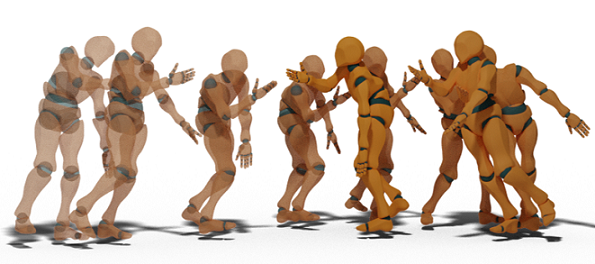}}}
\newcommand{\ssSty}{\raisebox{-.5\height}{\includegraphics[width=0.30\linewidth]{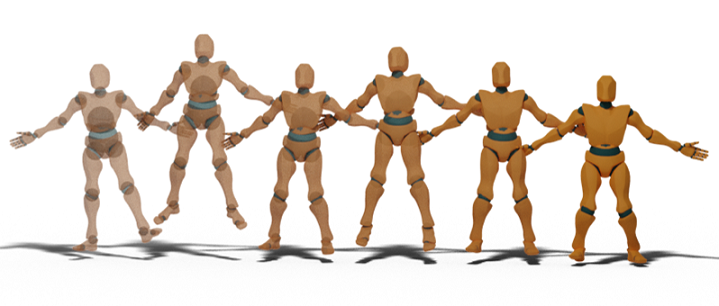}}}

\newcommand{\tCnt}{\raisebox{-.5\height}{\includegraphics[width=0.30\linewidth]{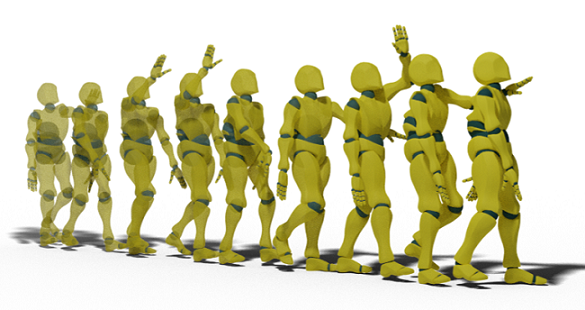}}}
\newcommand{\tfOurs}{\raisebox{-.5\height}{\includegraphics[width=0.30\linewidth]{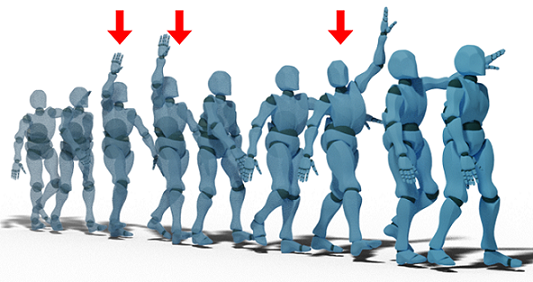}}}
\newcommand{\tsOurs}{\raisebox{-.5\height}{\includegraphics[width=0.30\linewidth]{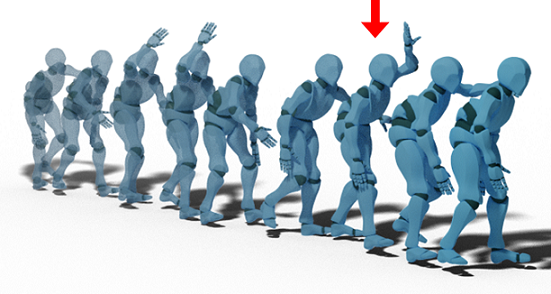}}}
\newcommand{\tfSty}{\raisebox{-.5\height}{\includegraphics[width=0.30\linewidth]{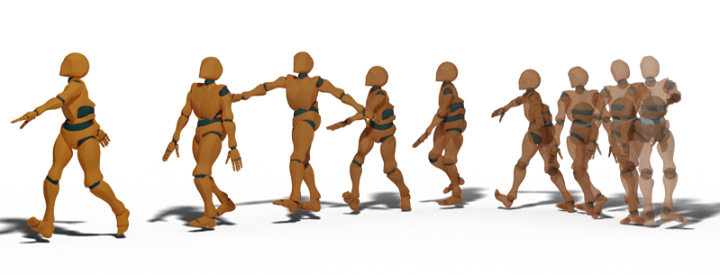}}}
\newcommand{\tsSty}{\raisebox{-.5\height}{\includegraphics[width=0.30\linewidth]{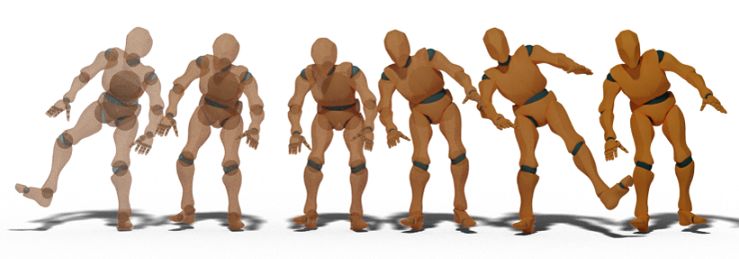}}}

\begin{figure*}[t]
\begin{minipage}{\linewidth}
    \centering
    \begin{tabular}{c|cc}
        \toprule
        \begin{tabular}{lr}  & \textit{Style Motion} \\&\\&\\&\\&\\ \textit{Content Motion} & \end{tabular} 
        & \begin{tabular}[c]{@{}c@{}}{Happy}\\\tfSty\end{tabular} 
        & \begin{tabular}[c]{@{}c@{}}{Old}\\\tsSty\end{tabular} 
        \\\midrule
        \begin{tabular}[c]{@{}c@{}} {Neutral} \\\tCnt\end{tabular} 
        & \tfOurs
        & \tsOurs
        \\\bottomrule
    \end{tabular}%
    \vspace{-0.3cm}
    \caption*{(a)}
\end{minipage}
\vspace{0.2cm}

\begin{minipage}{\linewidth}
    \centering
    \begin{tabular}{c|cc}
        \toprule
        \begin{tabular}{lr}  & \textit{Style Motion} \\&\\&\\&\\&\\ \textit{Content Motion} & \end{tabular} 
        & \begin{tabular}[c]{@{}c@{}}{Zombie}\\\fsSty\end{tabular} 
        & \begin{tabular}[c]{@{}c@{}}{Heavy}\\\ffSty\end{tabular} 
        \\\midrule
        \begin{tabular}[c]{@{}c@{}} {Angry} \\\fCnt\end{tabular} 
        & \fsOurs
        & \ffOurs
        \\\bottomrule
    \end{tabular}%
    \vspace{-0.3cm}
    \caption*{(b)}
\end{minipage}
\vspace{0.2cm}

\begin{minipage}{\linewidth}
    \centering
    \begin{tabular}{c|cc}
        \toprule
        \begin{tabular}{lr}  & \textit{Style Motion} \\&\\&\\&\\&\\ \textit{Content Motion} & \end{tabular} 
        & \begin{tabular}[c]{@{}c@{}}{Drunk}\\\sfSty\end{tabular} 
        & \begin{tabular}[c]{@{}c@{}}{Strutting}\\\ssSty\end{tabular} 
        \\\midrule
        \begin{tabular}[c]{@{}c@{}} {Female} \\\sCnt\end{tabular} 
        & \sfOurs
        & \ssOurs
        \\\bottomrule
    \end{tabular}%
    \vspace{-0.3cm}
    \caption*{(c)}
\end{minipage}
\caption{Qualitative results in the BFA~\cite{aberman2020unpaired} dataset. The BFA dataset comprises long motion sequences not segmented by contents. Therefore, we label only the style categories at the upper side of each motion clip. 
Each generated motion successfully reflects the desired styles, as highlighted in the motion segments indicated by the red arrows.
(a) The character straightens its arms and opens its chest in the `Happy' style, whereas it bends its back and arms in the `Old' style.
(b) The character walks staggering with arms extended forward in the `Zombie' style, and the `Heavy' style expresses the weight when pressing down on the ground.
(c) The `Drunk' style expresses staggering motion, while the character in the `Strutting' style opens its chest and arms.
For clearer visualization, please refer to the attached video.} 
\label{fig:qual_bfa}
\end{figure*}

\end{document}